\documentclass[journal]{IEEEtran}

\usepackage{amssymb}
\usepackage{amsmath}
\usepackage{amsfonts}
\usepackage{multirow}
\usepackage{graphicx}
\usepackage{textcomp}
\usepackage{amsthm}
\renewcommand{\vec}[1]{\boldsymbol{#1}}
\usepackage{setspace}
\usepackage{subfigure}
\usepackage{algorithmic}
\usepackage[ruled,vlined,boxed,linesnumbered]{algorithm2e}
\usepackage[dvipsnames]{xcolor}
\usepackage{bm}
\usepackage{url}
\usepackage{hyperref}
\usepackage{cite}

\allowdisplaybreaks[4]

\ifCLASSINFOpdf
\else
\fi

\begin{document}

\title{Look Closer to Your Enemy: Learning to Attack via Teacher-Student Mimicking}

\author{Mingjie Wang,
	Jianxiong Guo,~\IEEEmembership{Member,~IEEE},
        Sirui Li, 
        Dingwen Xiao,
	and Zhiqing Tang,~\IEEEmembership{Member,~IEEE}
	\thanks{
    Mingjie Wang, Sirui Li, and Dingwen Xiao are with the Guangdong Key Lab of AI and Multi-Modal Data Processing, Department of Computer Science, BNU-HKBU United International College, Zhuhai 519087, China. (e-mail: mjwang0606@gmail.com; sil089@ucsd.edu; dxiaoaf@connect.ust.hk)

    Jianxiong Guo is with the Advanced Institute of Natural Sciences, Beijing Normal University, Zhuhai 519087, China, and also with the Guangdong Key Lab of AI and Multi-Modal Data Processing, BNU-HKBU United International College, Zhuhai 519087, China. (e-mail: jianxiongguo@bnu.edu.cn)

    Zhiqing Tang is with the Advanced Institute of Natural Sciences, Beijing Normal University, Zhuhai 519087, China. (e-mail: zhiqingtang@bnu.edu.cn)
		
	\textit{(Corresponding author: Jianxiong Guo; Zhiqing Tang.)}
	}
	\thanks{Manuscript received April xxxx; revised August xxxx.}}

\markboth{Journal of \LaTeX\ Class Files,~Vol.~xx, No.~xx, July~2023}%
{Shell \MakeLowercase{\textit{et al.}}: Bare Demo of IEEEtran.cls for IEEE Journals}


\maketitle

\begin{abstract}
Deep neural networks have significantly advanced person re-identification (ReID) applications in the realm of the industrial internet, yet they remain vulnerable.  Thus, it is crucial to study the robustness of ReID systems, as there are risks of adversaries using these vulnerabilities to compromise industrial surveillance systems. Current adversarial methods focus on generating attack samples using misclassification feedback from victim models (VMs), neglecting VM's cognitive processes. We seek to address this by producing authentic ReID attack instances through VM cognition decryption. This approach boasts advantages like better transferability to open-set ReID tests, easier VM misdirection, and enhanced creation of realistic and undetectable assault images. However, the task of deciphering the cognitive mechanism in VM is widely considered to be a formidable challenge. In this paper, we propose a novel inconspicuous and controllable ReID attack baseline, LCYE (\textit{\textbf{L}ook \textbf{C}loser to \textbf{Y}our \textbf{E}nemy}), to generate adversarial query images. Specifically, LCYE first distills VM's knowledge via teacher-student memory mimicking the proxy task. This knowledge prior serves as an unambiguous cryptographic token, encapsulating elements deemed indispensable and plausible by the VM, with the intent of facilitating precise adversarial misdirection. Further, benefiting from the multiple opposing task framework of LCYE, we investigate the interpretability and generalization of ReID models from the view of the adversarial attack, including cross-domain adaption, cross-model consensus, and online learning process. Extensive experiments on four ReID benchmarks show that our method outperforms other state-of-the-art attackers with a large margin in white-box, black-box, and target attacks. The source code can be found at \url{https://github.com/MingjieWang0606/LCYE-attack_reid}.
\end{abstract}

\begin{IEEEkeywords}
Adversarial Attack, GAN, ReID, Memory Module, Image Classification.
\end{IEEEkeywords}

\section{Introduction}\label{sec:introduction}
\IEEEPARstart{P}{erson} re-identification (ReID) \cite{memory,transreid, li2020infrared, wei2020co, wei2018glad, li2019multi} aims to associate images of a person across disjoint cameras. In recent years, ReID methods predicated on deep learning \cite{transreid} have not only achieved accuracy rates exceeding 90\% but also transcended human-level competence. Nonetheless, it has been discovered that deep ReID models display a high susceptibility to adversarial instances, implying that even slight alterations to the initial instances could lead the model to erroneous conclusions with a high degree of certainty \cite{misranking}. Within the context of practical applications, such a susceptibility could be exploited either malevolently (for example, enabling criminals to evade ReID-based surveillance systems) or benevolently (for instance, permitting individuals with sensitive identities to remain concealed within a database).

\begin{figure}
    \centering
    \includegraphics[width=\linewidth]{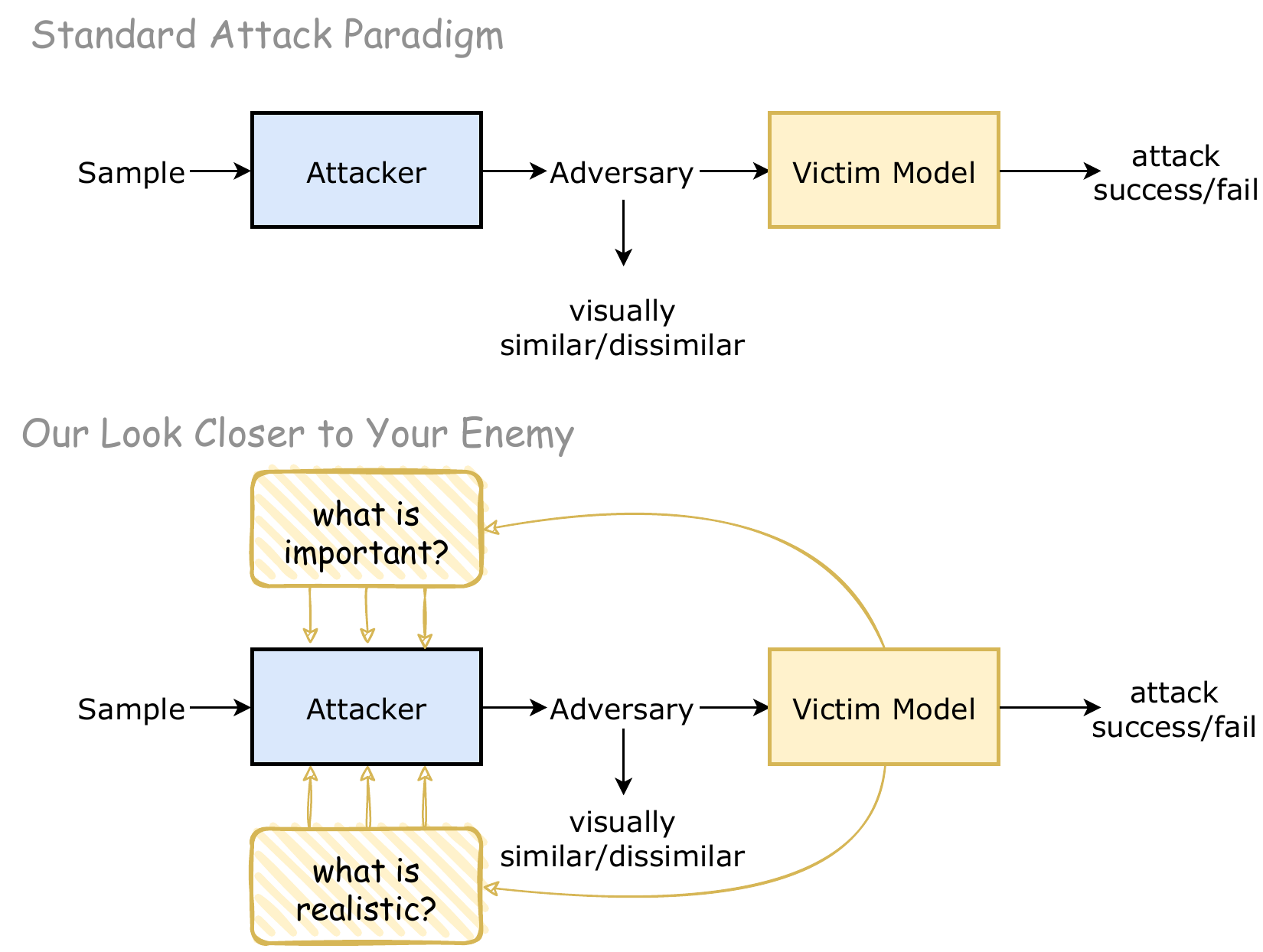}
    \caption{Illustration of standard attack paradigm and the idea of our \textit{Look Closer to Your Enemy.}}
    \label{fig1 paradigm}
\end{figure}

Current research endeavors attacking image retrieval models, as reflected in studies \cite{metricattack,misranking}, predominantly adopt paradigms borrowed from classification attacks. The standard attack paradigm as shown in Figure \ref{fig1 paradigm}, under the supervision for attack and imperceptible purposes, the attacker gradually optimizes toward a stable status between the attack performance and imperceptibility. However, we believe this paradigm is suboptimal. Firstly, it lacks pixel-level guidance since attack supervision gives global feedback for attack and imperception. Secondly, there exists a discrepancy in the training and attacking mechanisms for victim models (VMs). Specifically, ReID VMs are trained using varied structures and losses, such as PCB \cite{pcb} and circle loss \cite{circle}, yet they are attacked using a similar global mis-ranking loss \cite{misranking}. Lastly, the adversarial identity consistency for personalized and realistic generation remains unverified, both in RGB and latent space, particularly in Generative Adversarial Network (GAN)-based attackers \cite{misranking}. In such attackers, the discriminator distinguishes between real and fake, without indicating an association with the intended identity.

To handle the problems above, we propose to manipulate the perceptual framework of the VM to mount the attack. Specifically, the attacker may eliminate regions that the VM identifies as reflective of the genuine identity, while concurrently introducing elements associated with a different identity. A salient example of this can be seen in the attribute-level descriptions provided by datasets such as Market1501 \cite{market} and DukeMTMC \cite{duke}, which might state \textit{``a young man wears a yellow shirt, black pants, and a hat, and carries no bag''}. These attributes provide concrete instantiations of an individual. We can execute an attack that alters specific attributes, such as replacing a yellow shirt with green short sleeves. Simultaneously, it is imperative to uphold the logical coherence of the attack, for instance, by avoiding contradictory modifications like forcing an individual to appear as if they are wearing both trousers and shorts concurrently. This particular form of attack possesses three inherent advantages: 
\begin{itemize}
    \item The attacker could be transferable to a cross-domain test set due to the transferability of well-posed VM knowledge.
    \item It provides pixel-level guidance taking into account both realistic and harmful generation for the attack.
    \item By manipulating the VM's perceptual framework, it mitigates the need for subverting the VM during training to garner adequate negative feedback.
\end{itemize}

\begin{figure}[!t]
    \centering
    \includegraphics[width=\linewidth]{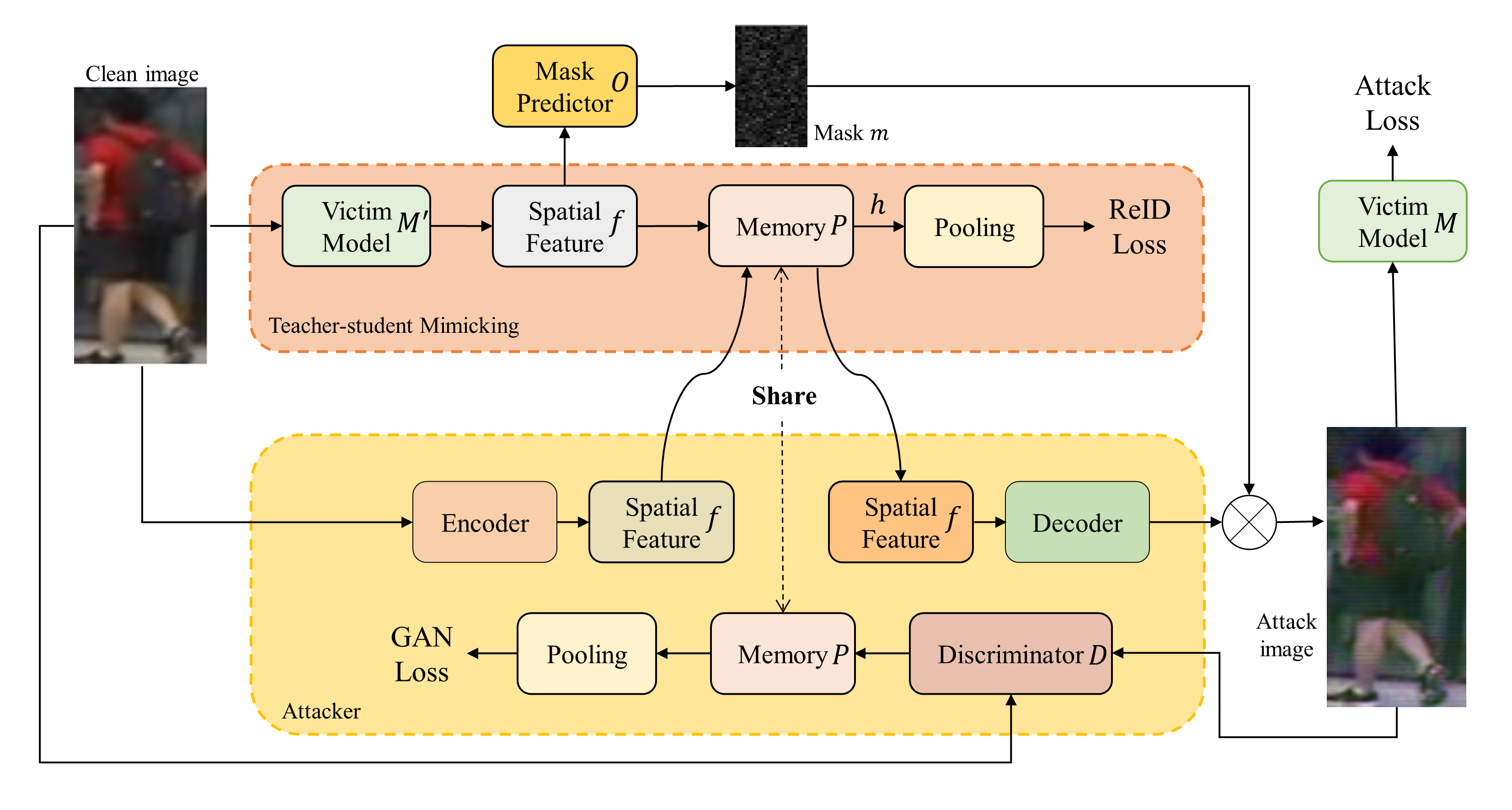}
   \caption{The unrolled framework of our \textit{Look Closer to Your Enemy} (LCYE). The generator of the attacker has an encoder and a decoder which retrieves relevant prototypes from the memory module at the middle of the encoder-decoder.}
   \label{fig1}
\end{figure}

Driven by this insight, we propose not to seek an instance-level accurate image, \textit{i.e.,} coercing one image to morph into another, but rather to search for a generalized identity-level structural prototype. In this paper, we introduce a novel, subtle, and controllable attack baseline harnessing the identity-level understanding of VMs. As depicted in Figure \ref{fig1}, we first capture the particular attribute of the VM using our proposed attention-like memory module via a teacher-student mimicry approach. Subsequently, we enable the attacker to retrieve the potential target prototype from the memory module. In specific terms, the knowledge of what the VM interprets as an identity is learned online and stored within a parameterized memory module. Both the generator and the discriminator then retrieve relevant prototypes from this memory, but with differing objectives. The generator obtains the prototype of potential targets for the attack, while the identity-wise discriminator verifies whether the generated image maintains the prototype of the targeted identity. We integrate the pixel-level attack and perceptibility measurement within a preservation-consistent Generative Adversarial Network framework.

An additional contribution of this paper is the exploration of the interpretability and generalization capacities of Recognition using IDentity (ReID) models from an adversarial attack perspective, encompassing cross-domain adaptation, cross-model consensus, and online learning processes. The discrete teacher and student models within the mimicking branch facilitate the analysis of explainable ReID across various hybrid combinations. In summary, our contributions can be outlined as follows:
\begin{itemize}
    \item We propose a novel ReID attack baseline named \textit{Look Closer to Your Enemy} (LCYE) to solve the issues from joint attack and imperceptibility optimization.
    \item We delve deep into interpretable and generalizable ReID from the aspect of adversarial attack by evaluating model consensus, covering cross-model and cross-domain covariance, and the online learning process.
    \item Our method obtains a promising attack success rate with inconspicuous noise. Experimental validations on four of the most extensive ReID benchmarks with both CNN and Transformer-based ReID SOTAs \cite{transreid,pcb} validate our method's superior efficiency and transferability in white-box, black-box, and target attacks.
\end{itemize}

\textbf{Organization: } In Section \ref{sec2},  we discuss the related work of Person Reid and Adversarial Attack. In Section \ref{sec4}, we establish the basic settings of our attack algorithm, introduce our novel ReID attack baseline, LCYE (Look Closer to Your Enemy), and its method of generating adversarial query images, and discuss the advantages of our approach and the challenges in deciphering VM's cognitive mechanism. Section \ref{sec5} details the experiments carried out on four ReID benchmarks, comparing our method against other state-of-the-art attackers. Lastly, Section \ref{sec6} wraps up the study and mentions potential directions for future research.

\section{Related Work}\label{sec2}
In this section, we discuss the related work of the realm of person re-identification (ReID) and Adversarial Attacks. ReID refers to the task of matching images of the same person across different camera views or time instances, typically used in surveillance and security applications. With the rise of deep learning techniques, ReID systems have shown significant improvements. However, they remain vulnerable to Adversarial Attacks. An Adversarial Attack introduces small, often imperceptible perturbations to the input data, designed to mislead a trained model into making a false classification or decision. By integrating these attacks into ReID, it becomes crucial to understand their implications and devise potential defense mechanisms to ensure the robustness of ReID systems.

\textbf{Person Re-identification} aims to spot the appearance of the same person in different observations \cite{circle}. Deep feature-based methods \cite{densenet,dong2018boosting} and metric learning-based methods \cite{rerank,chen2020deep} have achieved significant progress in supervised ReID. For deep feature-based methods, a cascade structure has already been studied in the neural network literature in the 1980s \cite{Cascade-Correlation}. Although fully connected cascade networks trained with batch gradient descent \cite{wilamowski2010neural} are effective on small datasets. This method only applies to networks with a few hundred parameters. In \cite{ sermanet2013pedestrian, yang2015multi}, it has been found to be effective for various vision tasks to utilize multi-level features in CNNs through skip-connections. Then, the pure theoretical framework of networks with cross-layer connections is derived \cite{cortes2017adanet}.
Highway Networks was the first batch of architectures that provided effective training methods for more than 100 layers of end-to-end networks \cite{srivastava2015training}. 

Additionally, ResNets have achieved impressive performance like ImageNet and COCO object detection \cite{resnet}. ResNets with pre-activation can also help train state-of-the-art networks with more than 1000 layers \cite{he2016identity}.
For metric learning-based methods, TriNet \cite{harwood2017smart} samples the most negative samples in the batch to achieve rapid convergence. The negative samples were found by Harwood et al. \cite{hermans2017defense} from the increasing search space defined by the nearest neighbor distance.
For example, TransReID \cite{transreid} achieves state-of-the-art performance,  with the participation of a local-aware Transformer \cite{vit}. Moreover, other innovative works \cite{liao2020interpretable,song2019generalizable} already focus on the interpretable and generalizable ReID which is verified by improved results. However, these methods may not be suitable and scalable for different tasks and models. In this paper, we aim to jointly analyze different ReID models in the point of supervised and unsupervised adversarial attack, keeping the fairness and flexibility of assessing the robustness and generalization. This attempt is not limited to cross-validation via simple white-box, black-box attacks but covers implicit decision-making.

\begin{table}[!t]
\renewcommand{\arraystretch}{1.1}
\caption{Key terminologies and concepts for LCYE}
\label{table}
\resizebox{\linewidth}{!}{
\begin{tabular}{|l|l|}
\hline
Symbol& 
Statement\\
\hline
$x$& 
original image \\
$x'$& 
adversarial counterpart of the original image\\
$N$& 
the identity number of the training set\\
$H , W$& 
height and width of the image \\
$f_i \in f$& 
the feature from the input\\
$k_j \in K$& 
the slices from prototype the matrix\\
$w_{ij}$& 
the normalized weight \\
$\mathcal{I}$& 
the clean image \\
$\mathcal{\hat{I}}$& 
the adversary of the clean image \\
$C_n$& 
the number of samples drawn from the n-th person ID \\
$\mathcal{I}^n_c$& 
the $c$-th images of the $n$ ID in a mini-batch \\
$c_s$& 
the samples from the same ID \\
$c_d$& 
the samples from the different ID \\
$\mathcal{L}_{*}$& 
the loss of  $*$ \\
$\lambda$, $\alpha_*$, $\beta_*$& 
the tradeoff factors \\

$\mathcal{G}$& 
the generator of the attacker \\
$\mathcal{P}$& 
memory module \\
$\mathcal{M}$& 
the misclassification model \\
$\mathcal{H}$& 
the pixel-level perception loss \\
$\mathcal{A}$& 
the global-level attack loss \\
$\partial L$& 
the joint gradient \\
$\mathcal{M}^{\prime}$& 
the subnet of VM \\
$n$&
noise \\
$\mathcal{O}$&
mask predictor \\
$m$&
mask \\
$h_{attack}$&
the attack cues \\
$\mathcal{S}$&
log-softmax function \\
$\mathbb{H}$&
indicator function \\
\hline
\end{tabular}}
\label{AdversarialTerms}
\end{table}

\textbf{Adversarial Attack} is to extract samples from real data to fool the learning model and help evaluate the robustness of the target models \cite{chakraborty2018adversarial, zhang2018face, li2019universal, Zhang2021OHRE, xie2023adaptive}. The security problems of the current most advanced model \cite{sharif2016accessorize, eykholt2018robust} and more insights into the CNN mechanism \cite{goodfellow2014explaining} was raised by Szegedy et al. The fast-gradient sign method \cite{goodfellow2014explaining}, which generates adversarial examples in one step, is one of the earliest works of gradient-based attack. The primary iterative method \cite{kurakin2018adversarial}, deep fool \cite{moosavi2016deepfool} and iterative momentum method \cite{dong2018boosting} extend the fast-gradient sign method \cite{goodfellow2014explaining} to update the adversarial images with small step sizes iteratively. Score-based attacks rely on searching input space. Single-pixel perturbation out of the valid image range can successfully lead to misclassification on small-scale images \cite{narodytska2016simple}, which can be extended to large-scale images by local greedy searching. In addition to pixel modification, the adversarial examples can also result in sample generation via spatial transform \cite{xiao2018spatially}. The iterative least-likely class method \cite{kurakin2018adversarial} increases the prediction probability of the least possible class through constraints, so the classification model outputs interesting errors. 

Adversarial Attack on ReID needs to generalize to unseen query images using misranking loss, which differs from adopting misclassification loss in close-set recognition tasks \cite{zhang2022close,gao2020dcr}. For optimization-based methods,  \cite{query} proposes an ODFA that exploits feature-level adversarial gradients. \cite{metricattack} proposes a metric attack to distort the distance between the attacked image and other similar images. Besides, \cite{misranking} introduces a GAN-based misranking attacker whose transferability is proved by its post-hoc results. However, these methods still ignore pixel-level supervision considering both attack and perception, which is essential to tackle their tradeoff. In this paper, we propose our LCYE to solve this issue by cheating VM's mind to generate harmful but realistic images.


\section{Methodology}\label{sec4}
In this section, we elucidate the challenges and methods associated with adversarial attacks. After detailing the mechanisms behind clean image misclassification, we discuss the constraints of close-set recognition tasks and the intricacies of GAN-based attackers. Notably, the conventional paradigm of adversarial attacks is contrasted with the adversarial reinforcement approach from VM. To address these complexities, we introduce a novel method named LCYE, designed to deceive VM's sample-driven decision-making process. The frequently used notations are summarized in Table \ref{AdversarialTerms}.

\subsection{Overall Framework}
\label{sec4:of}
The unrolled framework of our method is illustrated in Figure \ref{fig1}. We facilitate the learning of adversarial the attacker by jointly solving two opposing tasks: (1) memorizing the underlying recognition cues for ReID of each identity via teacher-student memory mimicking and (2) interpolating this prior knowledge of VM to the attacker to guide the efficient adversarial generation.

The objective of an adversarial attack, given a clean image and its ground truth label $(\vec{x}, y)$, is to lead the model $\mathcal{M}$ into making a misclassification on the input $\vec{x}$. The adversary $\vec{x}^{\prime}$ is determined based on:
\begin{equation}\label{eq:adversarial_objective}
 \min_{\vec{x}^{\prime}} \mathcal{H}\left(\vec{x}, \vec{x}^{\prime}\right) \text{and} \min_{\vec{x}^{\prime}}\mathcal{A}(\mathcal{M}\left(\vec{x}^{\prime}\right), y),
\end{equation}
where $\mathcal{H}$ represents the pixel-level perception loss and $\mathcal{A}$ signifies the global-level attack loss. Specifically, Pixel-level perception loss, denoted by $\mathcal{H}$, measures the perceptual difference between the original image $\vec{x}$ and its adversarial counterpart $\vec{x}^{\prime}$. The main purpose of this loss is to ensure that the adversarial image remains visually similar to the original image. Mathematically, this can be represented as:
\begin{equation}\label{eq2}
\mathcal{H}\left(\vec{x}, \vec{x}^{\prime}\right) = \frac{1}{H \times W} \sum_{i=1}^{H} \sum_{j=1}^{W} \|x_{i,j} - x^{\prime}_{i,j}\|^2,
\end{equation}
where $H$ and $W$ are the height and width of the image, respectively, and $\|\cdot\|$ denotes the L2 norm. By minimizing this loss, the perturbations added to the original image are kept subtle and almost imperceptible to human eyes. 
On the other hand, the global-level attack loss, denoted by $\mathcal{A}$, focuses on the task of misleading the model $\mathcal{M}$ into making incorrect predictions. Specifically, the loss ensures that the adversarial example $\vec{x}^{\prime}$ is classified differently than the original label $y$ of image $\vec{x}$. Formally, it can be expressed as:
\begin{equation}\label{eq3}
\mathcal{A}(\mathcal{M}\left(\vec{x}^{\prime}\right), y) = - \log(\mathbb{P}_{\mathcal{M}}(y|\vec{x}^{\prime})),
\end{equation}
where $\mathbb{P}_{\mathcal{M}}(y|\vec{x}^{\prime})$ indicates the model's predicted probability of the adversarial image $\vec{x}^{\prime}$ belonging to the correct class $y$. By maximizing this loss, we can ensure that the adversarial image is classified incorrectly by the model.

In essence, the balance between these two losses is crucial. While $\mathcal{H}$ ensures that the adversarial image remains visually close to the original, $\mathcal{A}$ guarantees that the model is effectively fooled by the perturbed image. Therefore, to avoid optimization collapse in solving such multiple opposing tasks, we deliberately separate mimicking and attack branches as much as possible. In particular, the clean images $\vec{x}$ are sent to the mimicking branch composed by the subnet $\mathcal{M}^{\prime}$ of VM, memory module $\mathcal{P}$, to learn identity-wise structural prototypes, which will be described in Section \ref{TeacherStudentMemoryMimicking}.

Afterward, the generator $\mathcal{G}$ of the attacker, composed of an encoder and a decoder, retrieves a similar but different identity prototype to generate, which determines to mislead the VM in prior. Meanwhile, according to the searchable realistic memory, the discriminator $\mathcal{D}$ distinguishes the adversarial generation belonging to the identity it claimed, following consistent identity preservation. This parts will be described in Section \ref{Memory-guidedAdversarialAttacker}.

\subsection{Teacher-student Memory Mimicking}
\label{TeacherStudentMemoryMimicking}
Inspired by \cite{memory}, we insert an external-attention memory module $\mathcal{P}$ into the VM to remember such prototypes in a learnable tensor cache. Specifically, we keep and freeze the subnet $\mathcal{M}^{\prime}$ of VM before pooling, local split, or equivalents to obtain informative spatial features where we can access the sampled-driven decision-making process. Then, for the following evaluation fairness, we apply the identical memory module $\mathcal{P}$,  MaxPooling, Batch Normalization (BN), and classification head sequentially after $\mathcal{M}^{\prime}$ for different VMs. During fine-tuning for ReID, $\mathcal{P}$ could dynamically record what VM recognizes images for each identity and what are real images embedded in latent space, \textit{i.e.,} identity-wise structural prototype. 

The prototype memory module $\mathcal{P}$ contains $N$ learnable identity prototypes, which are recorded by a matrix $\vec{K} \in \mathbb{R}^{N \times C}$ with fixed feature dimension $C$.  Identity prototype number $N$ equals the identity number of the training set where each identity contains one prototype in memory. Then an attention-based addressing operator for accessing the memory, \textit{i.e.}, memory reader and writer, is used to assign each image into spare prototypes. The role of the memory reader and writer depends on whether $\vec{K}$ updates in the current step. Given the output $\vec{f} \in \mathbb{R}^{H \times W \times C}$ of $\mathcal{M}^{\prime}$ in which $H$ and $W$ denote the height and width resolution, our memory module could be denoted  as:
\begin{equation}\label{eq5}
    w_{i j}=\frac{\exp \left(d\left(\vec{f}_{i}, \vec{k}_{j}\right)\right)}{\sum_{j=1}^{N} \exp \left(d\left(\vec{f}_{i}, \vec{k}_{j}\right)\right)}
\end{equation}
where $\vec{f}_{i}\in\mathbb{R}^C$ and $\vec{k}_{j}\in\mathbb{R}^C$ are feature and prototype slices from input $\vec{f}$ and prototype matrix $K$. $w_{ij}$ is the normalized weight measuring the cosine similarity $d(\cdot,\cdot)$ between $\vec{f}_{i}$ and $\vec{k}_{j}$. Thus, the assigned prototype $\vec{h}\in \mathbb{R}^{H\times W \times C}$ from feature $\vec{f}$ could be calculated as:
\begin{equation}\label{eq6}
    \vec{h} = MRA(\vec{f},K) = \oplus_{i=1}^{H\times W} \sum\nolimits_{j=1}^{N} w_{ij} \vec{k}_{j},
\end{equation}
where the $\oplus$ indicates that there are $H\times W$ vectors concatenating together. Thus, the dimension of $\vec{h}$ is $\mathbb{R}^{H\times W \times C}$, which is aligned with $\vec{x}$.

\subsection{Memory-guided Adversarial Attacker}
\label{Memory-guidedAdversarialAttacker}

The motivation of our LCYE is to build a prototype-consistent generation for an adversarial attack: The generator accords to one potential identity prototype to generate the adversary, where the selected prototype claims the identity belonging to the adversary. Then the discriminator could also tell whether the adversary belongs to the claimed identity by checking the prototype. Thus, our memory-guided attacker obtains three main components: adversarial generator $\mathcal{G}$ to generate noise $\vec{n}$, identity-wise multi-stage discriminator $\mathcal{D}$ to ensure identity preservation consistency, and a mask predictor $\mathcal{O}$ to estimate the effective noise mask $\vec{m}$. The final attack image is composed of   $\vec{x}'= \vec{m}\odot\vec{n} + \vec{x}$. Besides, VM's knowledge learned and recorded in $\vec{K}$, is interpolated into the generator and discriminator to facilitate the inconspicuous adversarial generation. Meanwhile, the mask predictor selects the location of recognition cues from VM for the accurate hit.  Given the real image $\vec{x}$, our generator first encode it to $\vec{f}$ and let it retrieve the target prototype from $\mathcal{P}$ as attack cues $\vec{h}_{attack}$  to generate noise $\vec{n}$:
\begin{align}
    &\vec{h}_{attack} = MRA(\vec{f},\vec{K}),\\
    &n = \mathcal{G}(x, h_{attack}),
\end{align}
Hence, mask predictor $\mathcal{O}$ maps the location of discriminative region to image resolution as mask $\vec{m}$ according to the clues of spatial features $\vec{f}$ from VM:
\begin{equation}
\vec{f} = \mathcal{M}'(\vec{x}), \vec{m}  = \mathcal{O}(\vec{f}).
\end{equation}
Finally, the multi-stage identity-aware discriminator classifies whether attack images $\vec{x}'$ and original images $\vec{x}$ meet the prototype belonging to the claimed identity. In particular, three subnetworks, receiving \{1, 1/4, 1/16\} areas of the original images as the input, are introduced in $\mathcal{D}$ to obtain a multi-scale response. Then, by pyramiding the features of different discriminator levels as \cite{misranking}, a series of downsampled results with a ratio of \{1/32, 1/16, 1/8, 1/4\} of the image is thus formulated for final prediction. We empirically let features $s_{1/4}$ with a 1/4 resolution ratio to retrieve the relevant prototype from $\vec{K}$ to check whether the image meets the imagination of VM. Solely interpolating VM knowledge to the discriminator brings implicit external-internal semantic consistency but still lacks explicit identity consistency for personalized supervision. Therefore, $\mathcal{D}$ is designed to estimate multi-identity probability $\vec{p} \in \mathbb{R}^{N+1} $, where $N$ is the total number of identities in the training set, and the additional dimension denotes the fake class. Thus, we have
\begin{equation}
\vec{p} = \mathcal{D}(\vec{x}, \vec{h}_d), \vec{h}_d = MRA(s_{1/4}, \vec{K}).
\end{equation}

Note that this prototype interpolation in the generator and discriminator seems to have no detailed map with one-hot ground truth since the learnable prototype matrix $\vec{K}$ records the general representation of each identity but does not know the correspondence. It means LCYE lacks explicit and strict prototype supervision for each identity and we can only hope the model learns such correspondence. However,  this risk is mitigated by the adversarial generative mechanism and static memory reading. The competitive generator and discriminator can not lasso the memory module since they can only read it. In particular, when the memory is meaningful for proxy recognition, the focus of competition thus moves to how to use the memory for better generation, rather than ignoring it.


\begin{figure*}[!t]
  \centering
  \subfigure[Cross-model Attack]{
		\includegraphics[width=0.32\textwidth]{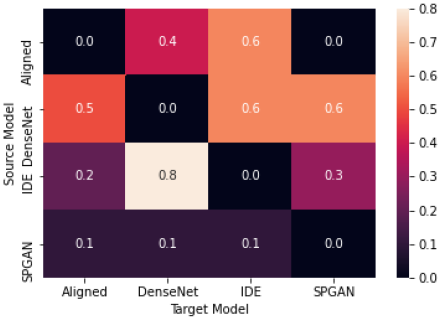}
  }%
  \subfigure[Cross-dataset Attack]{
		\includegraphics[width=0.32\textwidth]{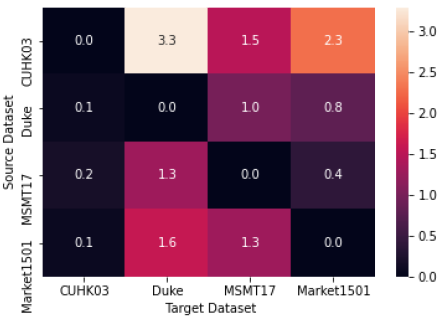}
  }%
  \subfigure[Cross-model \& Cross-dataset Attack]{
		\includegraphics[width=0.32\textwidth]{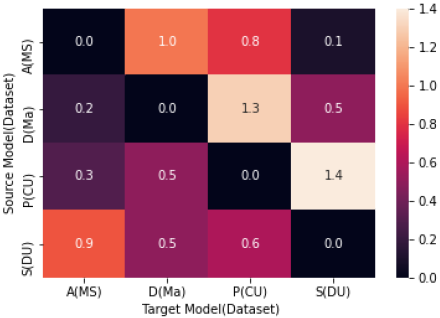}
  }%

  \subfigure[Cross-model Knowledge Distillation]{
		\includegraphics[width=0.32\textwidth]{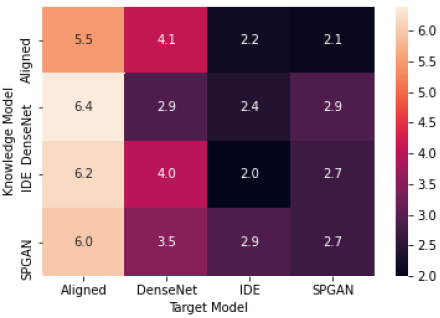}
  }%
  \subfigure[Cross-dataset Knowledge Distillation]{
		\includegraphics[width=0.32\textwidth]{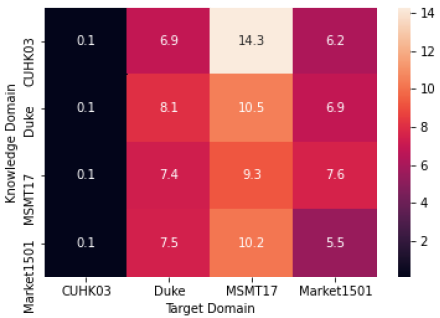}
  }%
  \subfigure[Cross-model \& Cross-dataset Knowledge Distillation]{
		\includegraphics[width=0.32\textwidth]{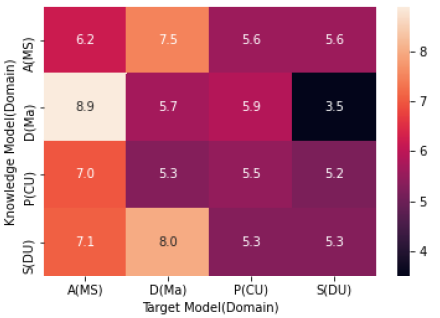}
  }%

  \caption{Rank1 results (\%) of \textit{black-box attack} and cross-model/domain knowledge distillation. In cross-modal\&dataset(domain) evaluation, we abbreviate AlignedReID(MSMT17)$\to$ A(MS), DenseNet(Market1501)$\to$D(Ma), PCB(CUHK03)$\to$P(CU), SPGAN(DukeMSMT)$\to$S(DU).}
  \label{figure blackbox}
\end{figure*}

\subsection{Objectives}
The mis-ranking loss, \textit{i.e.,} attack loss, includes four variants: (1) standard misclassification loss \textbf{cent}; (2) misclassification proposed by \cite{misranking} \textbf{xent}; (3) adversarial triplet loss \textbf{etri}; and (4) misclassification and adversarial triplet losses \textbf{xcent+etri}. We omit the description of standard misclassification loss \textbf{cent} since it is a simple adversarial cross-entropy loss. Given a clean image $\mathcal{I}$ and its adversary $\mathcal{\hat{I}}$,  the  misclassification loss \textbf{xent} proposed by \cite{misranking} is: 
\begin{equation}
\mathcal{L}_{x e n t}=-\sum_{n=1}^{N} \mathcal{S}(\mathcal{T}(\hat{\mathcal{I}}))_{n}\left((1-\delta) \mathbb{H}_{argmin \ \mathcal{T}(\mathcal{I})_{n}}+\delta v_{n}\right)
\end{equation}
where $\mathcal{S}$ is the log-softmax function, $N$ is the total identity number and $v=\left[\frac{1}{N-1}, \cdots, 0, \cdots, \frac{1}{N-1}\right]$ is smoothing regularization in which $v_k$ equals to $\frac{1}{N-1}$ everywhere except when $n$ is the ground-truth ID. $\mathbb{H}$ is the indicator function. The adversarial triplet loss \textbf{etri} is:
\begin{eqnarray}
	\mathcal{L}_{etri}&=& \sum_{n=1}^{N} \sum_{c=1}^{C_{n}}\left[\max_{{{j \neq n \atop j=1 \ldots N} \atop c_{d}=1 \ldots C_{j}} \atop \alpha_{j}}\left\|\mathcal{T}\left(\hat{\mathcal{I}}_{c}^{n}\right)-\mathcal{T}\left(\hat{\mathcal{I}}_{c_{d}}^{j}\right)\right\|_{2}^{2}\right.\nonumber\\
	~ &=& \left.-\min _{c_{s}=1 \ldots C_{n}}\left\|\mathcal{T}\left(\hat{\mathcal{I}}_{c}^{n}\right)-\mathcal{T}\left(\hat{\mathcal{I}}_{c_{s}}^{n}\right)\right\|_{2}^{2}+\Delta\right]_{+}
\end{eqnarray}
where $C_n$ is the number of samples drawn from the $n$-th person ID, $\mathcal{I}^n_c$ is the $c$-th images of the $n$ ID in a mini-batch,
$c_s$ and $c_d$ are the samples from the same ID and the different
IDs, $||\cdot||_2$  is the square of L2 norm used as the distance
metric, and $\Delta$ is a margin threshold.

For visual perception loss, we borrow two choices of \textbf{SSIM} and \textbf{MS-SSIM} from \cite{misranking}. The difference between them is interpolating multi-scale (\textbf{MS}) measurement. Therefore, we mainly represent \textbf{MS-SSIM} here:
\begin{align}
&\mathcal{L}_{MS-SSIM}(\mathcal{I}, \hat{\mathcal{I}})\nonumber\\
&=\left[l_{L}(\mathcal{I}, \hat{\mathcal{I}})\right]^{\alpha_{L}} \cdot \prod_{j=1}^{L}\left[c_{j}(\mathcal{I}, \hat{\mathcal{I}})\right]^{\beta_{j}}\left[s_{j}(\mathcal{I}, \hat{\mathcal{I}})\right]^{\gamma_{j}},
\end{align}
where $c_j$ and $s_j$ are the measures of the contrast comparison
and the structure comparison at the $j$-th scale respectively,
which are calculated by $c_{j}(\mathcal{I}, \hat{\mathcal{I}})=\frac{2 \sigma_{\mathcal{I}} \sigma_{\hat{\mathcal{I}}}+C_{2}}{\sigma_{\mathcal{I}}^{2}+\sigma_{\hat{\mathcal{I}}}^{2}+C_{2}}$ and $s_{j}(\mathcal{I}, \hat{\mathcal{I}})=\frac{\sigma_{\mathcal{I} \hat{\mathcal{I}}}+C_{3}}{\sigma_{\mathcal{I}} \sigma_{\mathcal{I}}+C_{3}}$ where $\sigma$ is the variance/covariance.

For GAN loss, we adopt the multi-discriminator and multi-label GAN loss as:
\begin{align}
    \mathcal{L}_{G A N}&=\mathbb{E}_{\left(I_{c d}, I_{c s}\right)}\left[\log \mathcal{D}_{1,2,3}\left(I_{c d}, I_{c s}\right)\right]\nonumber\\
    &+\mathbb{E}_{\mathcal{I}}\left[\log \left(1-\mathcal{D}_{1,2,3}(\mathcal{I}, \hat{\mathcal{I}})\right)\right],
\end{align}
where the subscript $1,2,3$ denotes our multi-stage discriminator. The identity-aware supervision could be deemed as the multi-label binary version of standard real/fake loss.

\subsection{Objective Function}
The total objective includes (1) mis-ranking loss $\mathcal{L}_{mr}$ for attack; (2) GAN loss $\mathcal{L}_{GAN}$ for personalized realistic generation; (3) visual perception loss $\mathcal{L}_{VP}$ for inconspicuous change; and (4) cross-entropy loss $\mathcal{L}_{ce}$ and triplet loss $\mathcal{L}_{tri}$ for teacher-student mimicking:
\begin{align}
    &\mathcal{L}_{attack} = \alpha_1 \cdot \mathcal{L}_{mr}+ \alpha_2 \cdot \mathcal{L}_{GAN} + \alpha_3 \cdot \mathcal{L}_{VP},\\
    &\mathcal{L}_{mimic} = \beta_1 \cdot \mathcal{L}_{ce} + \beta_2 \cdot \mathcal{L}_{tri},
\end{align}
where $\alpha_*$ and $\beta_*$ are tradeoff factors. $\mathcal{L}_{attack}$ is the loss for the attack branch, while $\mathcal{L}_{mimic}$ is for mimicking the branch. Since the objectives are already proposed or largely identical as Mis-ranking \cite{misranking} that we modify from, we prefer to only discuss the capability of inserting VM's knowledge in the attacker, which eases the dependence on both mis-ranking loss $\mathcal{L}_{mr}$ and visual perception loss $\mathcal{L}_{VP}$ in Section \ref{loss_sensity}.  Empirically, we set $\alpha_1 = 1, \alpha_2 = 1,\ \alpha_3 = 1,\ \beta_1=1,$ and $\beta_2 = 1$.

\begin{figure}[!t]
  \centering
  \subfigure[Clean images]{
		\includegraphics[width=0.97\linewidth]{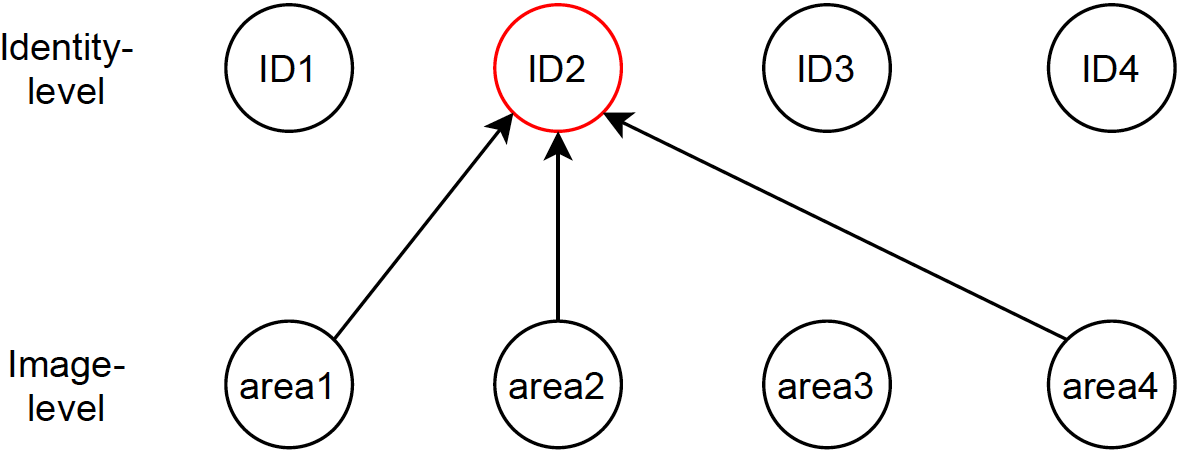}
  }%
  
  \subfigure[Adversary of current methods with standard paradigm]{
		\includegraphics[width=0.97\linewidth]{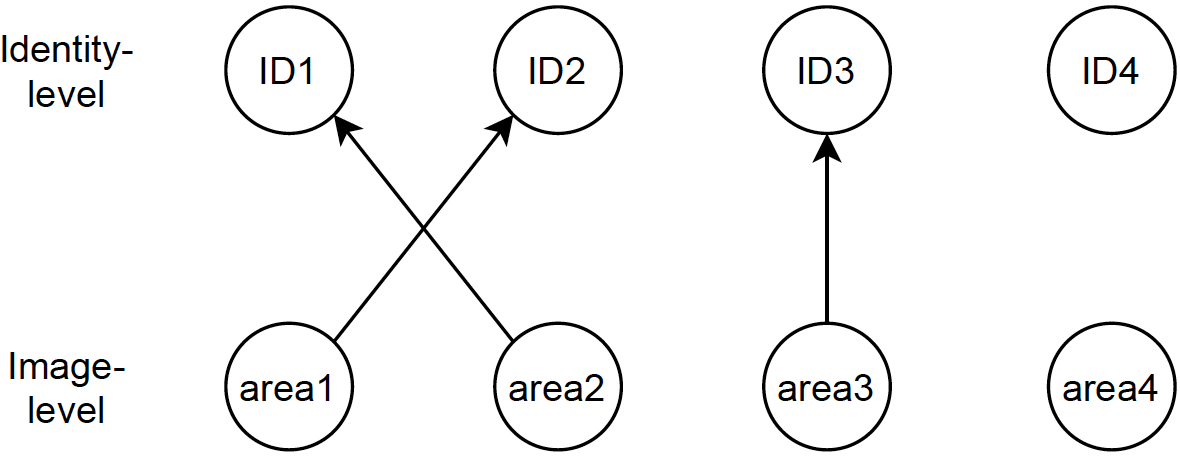}
  }%
  
  \subfigure[Adversary of our LCYE]{
		\includegraphics[width=0.97\linewidth]{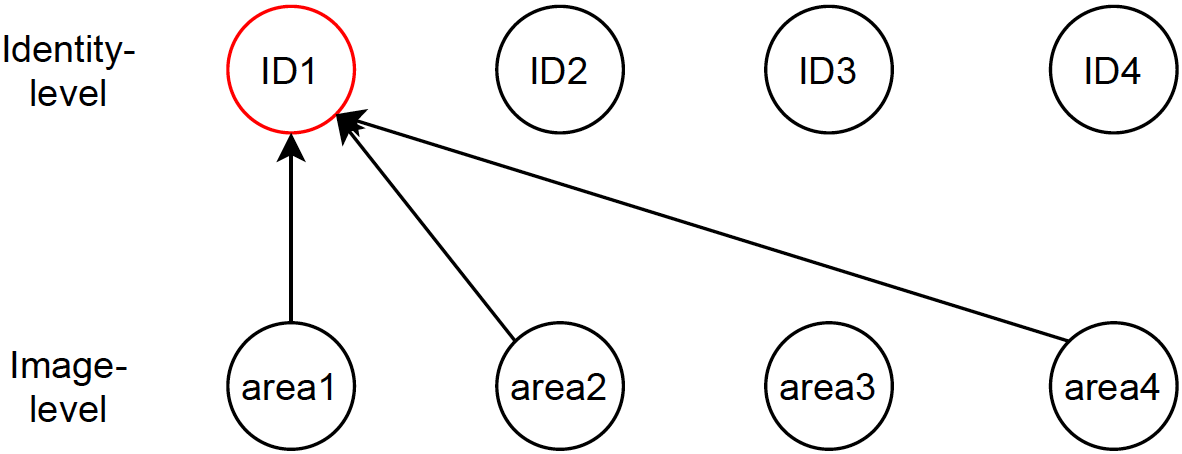}
  }%

  \caption{Illustration of decision making of victim model using clean images, adversarial images generated by current methods, and our LCYE. We omit the insignificant links from area to identity (ID).}
  \label{figure decision}
\end{figure}

\section{How Memory Help Attack?}\label{sec4-new}
To comprehend the underlying mechanics, we ought to revisit the foundational framework delineated in Sec. \ref{sec4:of}. Systematically, this can be partitioned into three integral components:

Firstly, the marriage of pixel-level perception loss and global-level attack loss appears incomplete, given that it does not offer pixel-level supervision that takes into account their combined objectives. The perception loss $\mathcal{H}$ and adversarial loss $\mathcal{A}$ operate in separate domains, probing the potential adversary $\vec{x}'$ in both RGB and latent spaces. The composite gradient $\partial L = \frac{\partial \mathcal{H}}{ \partial \vec{x}} + \lambda \cdot \frac{\partial \mathcal{A}}{\partial \vec{x}}$ is guided by the predetermined trade-off factor $\lambda$, which tends to confine the optimization process to a localized, biased optimum.

Secondly, in the realm of close-set recognition tasks—where training and test datasets have identical categories—Virtual Models (VM) $\mathcal{M}$s are educated using a cross-entropy loss and subsequently assaulted with an adversarial variant of the same. This seems intuitive. Nonetheless, when it comes to the open-set ReID models $\mathcal{M}$, their training procedures exhibit a wide variation in terms of structure and loss functions. As an illustration, while PCB aims to mine local representation, $\mathcal{M}$ finds itself under the assault of an adversarial triplet loss, which targets global features. This approach seems to neglect the unique attributes of individual models. Although tailoring attack supervision for each model could be a potential remedy, its implementation remains challenging.

Thirdly, the commitment to adversarial category consistency within GAN-based attackers \cite{misranking,hu2017generating} appears somewhat diluted. These methodologies typically employ a fixed VM as an outboard category discriminator while concurrently training an online real/fake discriminator to satiate the need for authentic generation. However, such a static external VM can be effortlessly deceived by the generator. Concurrently, the rudimentary real/fake discriminator struggles to discern genuinely individualized images from merely authentic ones. A deeper dive into this concern can be found in Section \ref{sec:category}, wherein we highlight that the images attacked via the GAN-based approach tend to manifest more apparent noise.

At its core, we posit that the prevailing adversarial attack framework resembles an adversarial reinforcement exercise steered by VM. The challenges we've pinpointed stem from a tendency to overlook learning from samples and understanding the VM's perspective. To address these intertwined issues, this paper introduces a groundbreaking method, dubbed LCYE, designed to deceive the sample-driven decision-making machinery of the VM. Specifically, 

\begin{enumerate}
    \item The pixel-level adversarial direction is derived from the collaboration of the mask predictor and the memory reading component within the generator. These mechanisms not only determine the spatial regions to target but also decipher the identity transformation in the latent space. In contrast, existing techniques \cite{misranking} either fall short in terms of precise control or predominantly rely on misclassification-based adversarial feedback.
    
    \item Addressing the unique characteristics of the victim model, our LCYE introduces a universal teacher-student mimicry approach. Instead of seeking an adversarial analogue, such as the hard adversarial triplet loss or the simple triplet loss, LCYE preserves the original architecture and protocols of the victim model, while embracing a multi-faceted adversarial strategy.
    
    \item Through prototype interpolation combined with identity-aware generation—termed as category cycle consistency—we ensure that every modified region directly contributes to the recognition of the desired identity. Figure \ref{figure decision} illustrates this distinction: whereas conventional paradigms distribute the significance of different regions across multiple identities, our approach coherently aligns these influential areas to a singular, erroneous higher-level concept by consistently associating the same identity prototype with the image.
\end{enumerate}

Concluding, based on the aforementioned tripartite decomposition, we can ascertain that the conventional methodologies embody certain limitations and biases when juxtaposed against the demands of real-world adversarial tasks. Our LCYE, conceived in response to these shortcomings, not only provides a holistic approach but also ensures precise and controllable adversarial perturbations. By profoundly understanding the VM's standpoint and meticulously addressing each of its intricacies, LCYE offers a robust mechanism to effectively operate under diverse adversarial conditions.

\begin{table*}[!t]
\renewcommand{\arraystretch}{1.2}
\caption{Attacking the state-of-the-art ReID systems on Market1501. \textcolor{blue}{blue} and \textcolor{red}{red} denote previous best and current best results. $\downarrow$ means the lower numerical value is better for attack.}
\label{table whitebox}
\centering
\resizebox{\linewidth}{!}{
\begin{tabular}{cc|cccc|cccc|cccc|cccc}
\hline
\multicolumn{2}{c|}{\multirow{2}{*}{Methods with Market1501}}                                & \multicolumn{4}{c|}{Rank1$\downarrow$} & \multicolumn{4}{c|}{Rank5$\downarrow$} & \multicolumn{4}{c|}{Rank10$\downarrow$} & \multicolumn{4}{c}{mAP$\downarrow$} \\ \cline{3-18} 
\multicolumn{2}{c|}{}                              & Before        & PGD        & MR        & Ours       & Before       & PGD        & MR         & Ours       & Before        & PGD        & MR         & Ours       & Before       & PGD        & MR       & Ours      \\ \hline
\multicolumn{1}{c|}{\multirow{3}{*}{Backbone}}     & IDE(ResNet50)           & 83.1          & 4.5        & \textcolor{blue}{3.7}       & \textcolor{red}{0.3}        & 91.7         & 8.7        & \textcolor{blue}{8.3}        & \textcolor{red}{1.2}        & 94.6          & 12.1       & \textcolor{blue}{11.5}       & \textcolor{red}{2.4}        & 63.3         & 4.6        & \textcolor{blue}{4.4}      & \textcolor{red}{0.3}       \\
\multicolumn{1}{c|}{}                              & DenseNet121             & 89.9          & \textcolor{blue}{1.2}        & \textcolor{blue}{1.2}       & \textcolor{red}{0}          & 96.0         & \textcolor{blue}{1.0}        & 1.3        & \textcolor{red}{0.2}        & 97.3          & \textcolor{blue}{1.5}        & 2.1        & \textcolor{red}{0.4}        & 73.7         & \textcolor{blue}{1.3}        & \textcolor{blue}{1.3}      & \textcolor{red}{0.2}       \\
\multicolumn{1}{c|}{}                              & Mudeep(Inceptionv3)     & 73.0          & 2.6        & \textcolor{blue}{1.7}       & \textcolor{red}{0.0}          & 90.1         & 5.5        & \textcolor{blue}{1.7}        & \textcolor{red}{0.2}        & 93.1          & 6.9        & \textcolor{blue}{5.0}        & \textcolor{red}{0.5}        & 49.9         & 2.0        & \textcolor{blue}{1.8}      & \textcolor{red}{0.2}       \\ \hline
\multicolumn{1}{c|}{\multirow{3}{*}{Part-Aligned}} & AlignedReID             & 91.8          & 10.2       & \textcolor{blue}{1.4}       & \textcolor{red}{0.1}        & 97.0         & 15.8       & \textcolor{blue}{3.7}        & \textcolor{red}{0.9}        & 98.1          & 19.1       & \textcolor{blue}{5.4}        & \textcolor{red}{1.8}        & 79.1         & 8.9        & \textcolor{blue}{2.3}      & \textcolor{red}{0.3}       \\
\multicolumn{1}{c|}{}                              & PCB                     & 88.6          & 6.1        & \textcolor{blue}{5.0}       & \textcolor{red}{0.0}          & 95.5         & 12.7       & \textcolor{blue}{10.7}       & \textcolor{red}{0.1}        & 97.3          & 15.8       & \textcolor{blue}{14.3}       & \textcolor{red}{0.2}        & 70.7         & 4.8        & \textcolor{blue}{4.3}      & \textcolor{red}{0.2}       \\
\multicolumn{1}{c|}{}                              & HACNN                   & 90.6          & 6.1        & \textcolor{red}{0.9}       & 2.8        & 95.9         & 8.8        & \textcolor{red}{1.4}        & 8.1        & 97.4          & 10.6       & \textcolor{red}{2.3}        & 12.9       & 75.3         & 5.3        & \textcolor{blue}{1.5}      & \textcolor{red}{1.2}       \\ \hline
\multicolumn{1}{c|}{\multirow{4}{*}{GAN}}          & CamStyle+Era(IDE)       & 86.6          & 15.4       & \textcolor{blue}{3.9}       & \textcolor{red}{0.1}        & 95.0         & 23.9       & \textcolor{blue}{7.5}        & \textcolor{red}{0.7}        & 96.6          & 29.1       & \textcolor{blue}{10.0}       & \textcolor{red}{1.4}        & 70.8         & 12.6       & \textcolor{blue}{4.2}      & \textcolor{red}{0.2}       \\
\multicolumn{1}{c|}{}                              & LSRO(DenseNet121)       & 89.9          & 7.2        & \textcolor{blue}{0.9}       & \textcolor{red}{0.8}        & 96.1         & 13.1       & \textcolor{blue}{2.2}        & \textcolor{red}{2.2}        & 97.4          & 15.2       & \textcolor{red}{3.1}        & 3.5        & 77.2         & 8.1        & \textcolor{red}{1.3}      & 1.7       \\
\multicolumn{1}{c|}{}                              & HHL(IDE)                & 82.3          & 5.7        & \textcolor{blue}{3.6}       & \textcolor{red}{0.1}        & 92.6         & 9.8        & \textcolor{blue}{7.3}        & \textcolor{red}{0.7}        & 95.4          & 12.2       & \textcolor{blue}{9.7}        & \textcolor{red}{1.4}        & 64.3         & 5.5        & \textcolor{blue}{4.1}      & \textcolor{red}{0.2}       \\
\multicolumn{1}{c|}{}                              & SPGAN(IDE)              & 84.3          & 10.1       & \textcolor{blue}{1.5}       & \textcolor{red}{0.0}          & 94.1         & 16.7       & \textcolor{blue}{3.1}        & \textcolor{red}{0.6}        & 96.4          & 20.9       & \textcolor{blue}{4.3}        & \textcolor{red}{1.6}        & 66.6         & 8.6        & \textcolor{blue}{1.6}      & \textcolor{red}{0.2}       \\ \hline
\multicolumn{1}{c|}{\multirow{2}{*}{Transformer}}  & TransReID(ViT+baseline) & 94.6              &  -          &    \textcolor{blue}{6.2}       &  \textcolor{red}{0.9}          &   98.2           & -           & \textcolor{blue}{10.0}           & \textcolor{red}{1.5}           & 99.2              &      -      &  \textcolor{blue}{12.1}          & \textcolor{red}{2.7}           & 87.1             & -           & \textcolor{blue}{6.3}         &      \textcolor{red}{0.8}     \\
\multicolumn{1}{c|}{}                              & TransReID(ViT)          &  95.1             &  -         &    \textcolor{blue}{5.2}       & \textcolor{red}{0.9}           &  98.4            & -           &  \textcolor{blue}{10.1}          &  \textcolor{red}{1.7}          &  99.1             &  -          &  \textcolor{blue}{12.0}          & \textcolor{red}{2.6}           &  89.0            &   -         &  \textcolor{blue}{6.5}        & \textcolor{red}{0.9}          \\ \hline
\end{tabular}}
\end{table*}

\begin{table*}[!t]
	\renewcommand{\arraystretch}{1.2}
	\caption{Attacking the state-of-the-art ReID systems on CUHK03.}
	\centering
	
	\resizebox{\linewidth}{!}{
		\begin{tabular}{cc|cccc|cccc|cccc|cccc}
			\hline
			\multicolumn{2}{c|}{\multirow{2}{*}{Methods with CUHK03}}                            & \multicolumn{4}{c|}{Rank1$\downarrow$} & \multicolumn{4}{c|}{Rank5$\downarrow$} & \multicolumn{4}{c|}{Rank10$\downarrow$} & \multicolumn{4}{c}{mAP$\downarrow$}   \\ \cline{3-18} 
			\multicolumn{2}{c|}{}                              & Before  & PGD & MS  & Ours & Before  & PGD & MS  & Ours & Before  & PGD  & MS  & Ours & Before & PGD & MS  & Ours \\ \hline
			\multicolumn{1}{c|}{\multirow{3}{*}{Backbone}}     & IDE(ResNet50)       & 24.9    & 0.8 & \textcolor{blue}{0.4} & \textcolor{red}{0.0}  & 43.3    & 1.2 & \textcolor{blue}{0.7} & \textcolor{red}{0.4}  & 51.8    & 2.1  & \textcolor{blue}{1.5} & \textcolor{red}{0.4}  & 24.5   & \textcolor{blue}{0.8} & 0.9 & \textcolor{red}{0.2}  \\
			\multicolumn{1}{c|}{}                              & DenseNet121         & 48.4    & 0.1 & \textcolor{blue}{0.0} & \textcolor{red}{0.0}  & 50.1    & \textcolor{red}{0.1} & 0.2 & 0.6  & 70.1    & \textcolor{red}{0.3}  & 0.6 & 1.2  & 84.0   & \textcolor{red}{0.2} & 0.3 & 0.4  \\
			\multicolumn{1}{c|}{}                              & Mudeep(Inceptionv3) & 32.1    & 0.4 & \textcolor{blue}{0.1} & \textcolor{red}{0.0}  & 53.3    & 1.0 & \textcolor{blue}{0.5} & \textcolor{red}{0.2}  & 64.1    & 1.5  & \textcolor{blue}{0.8} & \textcolor{red}{0.4}  & 30.1   & 0.8 & \textcolor{blue}{0.3} & \textcolor{red}{0.1}  \\ \hline
			\multicolumn{1}{c|}{\multirow{3}{*}{Part-Aligned}} & AlignedReID         & 61.5    & 1.4 & \textcolor{blue}{1.4} & \textcolor{red}{0.0}  & 79.4    & 2.2 & \textcolor{blue}{3.7} & \textcolor{red}{0.6}  & 85.5    & \textcolor{blue}{4.1}  & 5.4 & \textcolor{red}{1.1}  & 59.6   & \textcolor{blue}{2.1} & \textcolor{blue}{2.1} & \textcolor{red}{0.3}  \\
			\multicolumn{1}{c|}{}                              & PCB                 & 50.6    & 0.5 & \textcolor{blue}{0.2} & \textcolor{red}{0.0}  & 71.4    & 2.1 & \textcolor{blue}{1.3} & \textcolor{red}{0.2}  & 78.7    & 4.5  & \textcolor{blue}{1.8} & \textcolor{red}{0.8}  & 48.6   & 1.2 & \textcolor{blue}{0.8} & \textcolor{red}{0.3}  \\
			\multicolumn{1}{c|}{}                              & HACNN               & 48.0    & 0.4 & \textcolor{blue}{0.1} & \textcolor{red}{0.0}  & 69.0    & 0.9 & \textcolor{red}{0.3} & 0.4  & 78.1    & 1.3  & \textcolor{red}{0.4} & 1.1  & 47.6   & 0.8 & \textcolor{blue}{0.4} & \textcolor{red}{0.3}  \\ \hline
		\end{tabular}
	}
	\label{figure CUHK}
\end{table*}

\begin{table*}[!t]
	\renewcommand{\arraystretch}{1.2}
	\caption{Attacking the state-of-the-art ReID systems on DukeMTMC.}
	\centering
	
	\resizebox{\linewidth}{!}{
		\begin{tabular}{cc|cccc|cccc|cccc|cccc}
			\hline
			\multicolumn{2}{c|}{\multirow{2}{*}{Methods with DukeMTMC}}                       & \multicolumn{4}{c|}{Rank1$\downarrow$} & \multicolumn{4}{c|}{Rank5$\downarrow$} & \multicolumn{4}{c|}{Rank10$\downarrow$} & \multicolumn{4}{c}{mAP$\downarrow$}    \\ \cline{3-18} 
			\multicolumn{2}{c|}{}                                               & Before & PGD  & MR  & Ours & Before & PGD  & MR  & Ours & Before  & PGD  & MR  & Ours & Before & PGD  & MR  & Ours \\ \hline
			\multicolumn{1}{c|}{\multirow{4}{*}{GAN-based}} & CamStyle+Era(IDE) & 76.5   & 22.9 & \textcolor{blue}{1.2} & \textcolor{red}{0.6}  & 86.8   & 34.1 & \textcolor{blue}{2.6} & \textcolor{red}{1.5}  & 90.0    & 39.9 & \textcolor{blue}{3.4} & \textcolor{red}{2.6}  & 58.1   & 16.8 & \textcolor{blue}{1.5} & \textcolor{red}{0.3}\\ 
			\multicolumn{1}{c|}{}                           & LSRO(DenseNet121) & 72.0   & 7.2  & \textcolor{blue}{0.7} & \textcolor{red}{0.5}  & 85.7   & 12.5 & \textcolor{blue}{1.6} & \textcolor{red}{1.4}  & 89.5    & 18.4 & \textcolor{blue}{2.2} & \textcolor{red}{2.2}  & 55.2   & 8.1  & \textcolor{blue}{0.9} & \textcolor{red}{0.8}  \\ 
			\multicolumn{1}{c|}{}                           & HHL(IDE)          & 71.4   & 9.5  & \textcolor{blue}{1.0} & \textcolor{red}{0.1}  & 83.5   & 15.6 & \textcolor{blue}{2.0} & \textcolor{red}{0.8}  & 87.7    & 19.0 & \textcolor{blue}{2.5} & \textcolor{red}{1.7}  & 51.8   & 7.4  & \textcolor{blue}{1.3} & \textcolor{red}{0.2}  \\
			\multicolumn{1}{c|}{}                           & SPGAN(IDE)        & 73.6   & 12.4 & \textcolor{red}{0.1} & 0.4  & 85.2   & 21.1 & \textcolor{red}{0.5} & 1.2  & 88.9    & 26.3 & \textcolor{red}{0.6} & 2.5  & 54.6   & 10.2 & \textcolor{blue}{0.3} & \textcolor{red}{0.3}  \\ \hline
	\end{tabular}}
	\label{figure DUKE}
\end{table*}

\section{Experiments}\label{sec5}
In this section, we compare our proposed CLYE with other commonly used baselines.

\subsection{Datasets and Experimental Setup}
\textbf{Datasets.} Four ReID benchmarks are used for the evaluation of our method, including Market1501 \cite{market} (1,501 identities with 32,688 images), CUHK03 \cite{cuhk} (1,467 identities and 28,192 images), DukeMTMC \cite{duke} (1,404 identities with 34,183 images), and MSMT17 \cite{msmt} (4,101 identities and 126,441 images). For attack evaluation metrics, we use Rank1, 5, and 10, and mAP for ReID attack where the lower numerical value means a better success attack rate in an attack problem.

\textbf{Implementation Details.} We use batch size 32, learning rate 0.0002 for GAN, and 0.0003 for mimicking on a single GTX P40 GPU. We use \textbf{xent+etri} for misranking loss $\mathcal{L}_{mr}$ and \textbf{MS-SSIM} for visual perception loss $\mathcal{L}_{VP}$. The triplet margin is set to $0.3$ for mimicking and attacking branches. Compared with other attackers, we use full-size images as possible masks to attack. Our generator uses ResNet Block with $4\times$ downsampling and $4\times$ upsampling. The sub-discriminator adopts the basic structure of Mis-ranking. For target attacks, since ReID is an open-set task where training and test sets have non-overlapping identities, it is unfeasible to follow a close-set target attack. We separate it into two evaluations: target consistency on adversarial query images and standard attack. For each identity $n\in N$, we randomly select $\gamma$ query images for each identity to generate adversarial query images with a total number $\gamma \cdot N$. Then we sent these images to the victim model and calculated the Euclidean distance of their embeddings. With pseudo labels, we calculate the Rank1 accuracy as their target consistency.  

\textbf{Implementation Details and Protocols.} Our victim models include CNN-based methods (AlignedReID \cite{alignedreid}, DenseNet \cite{densenet}, etc \cite{ide,mudeep,hacnn}), GAN-based methods (CamStyle \cite{camstyle}, SPGAN \cite{spgan}, LSRO\cite{lsro}, HHL\cite{hhl}) and Transformer-based methods (TransReID \cite{transreid}). The basic framework of our LCYE largely benefited from Mis-ranking \cite{misranking}, including the hyper-parameters, basic model structure etc. Therefore, we compare it in ablation to verify the capability brought from VM knowledge. For a fair comparison, we adopt the same protocols as \cite{misranking} by $L_{\infty}$-bounded attacks with $\varepsilon=16$. The \textit{black-box attack} includes a cross-model attack, cross-dataset attack, and cross-model-dataset attack as the standard setting.   
For the \textit{target attack}, we achieve it by modifying Eqn. (\ref{eq6}) to $\oplus_{i=1}^{H\times W} \sum_{j=1}^{N}  w_{ij} \vec{q}_j \vec{k}_j$ where $\vec{q}_j \in Q\in \mathbb{R}^{N\times1}$ is the indicator of selected identity and adding corresponding identity consistency supervision in GAN. Due to the open-set essence of the ReID task, we evaluate the target attack performance by (1) the similarity matrix of embeddings of adversarial query images which is simplified as Rank1$\uparrow$ as target consistency and (2) the attack success rate, \textit{i.e.,} Rank1$\downarrow$ and mAP$\downarrow$.

For the ablation study, we clarify that (1) \textit{baseline mimicking} is to let $\mathcal{M}^{\prime}$ fixed and learn the memory; (2) \textit{online mimicking} is to allow $\mathcal{M}^{\prime}$ pre-trained on ImageNet to update with mimicking branch (learn from scratch as a ReID task); and (3) \textit{offline mimicking} is to first train to mimic branch with fixed $\mathcal{M}^{\prime}$ and then train the pending attack branch. If without specific instruction, all ablation experiments are done on AlignedReID with Market1501.

\subsection{Attacking State-of-the-Art ReID Models}
\label{sec:category}
\textbf{White-box Attack.} As shown in Table \ref{table whitebox}, Table \ref{figure CUHK} and Table \ref{figure DUKE}, we demonstrate the superior performance of our method against other attackers. Results on multiple datasets and models show that our LCYE essentially improves the attack success scope by 1\% for Rank1, 3\% for Rank5, 4\% for Rank10, and 2\% for mAP, respectively, in most cases. Moreover, our method performs favorably against Mis-ranking (MR) and PGD \cite{pgd} with a large margin on different lines of ReID models.

\textbf{Black-box Attack.} Figure \ref{figure blackbox} shows the cross-model \& cross-dataset \& cross-model-dataset evaluation. Benefiting from cheating VM's knowledge, our method achieves a similar attack performance as a white-box attack. We find that training on domain adaption SPGAN seems more versatile than others in the cross-model case, and results on CUHK03 in the cross-dataset evaluation show the domain-specific vulnerability.

\begin{figure*}[!t]
  \centering
  \subfigure[Original images]{
		\includegraphics[width=0.32\textwidth]{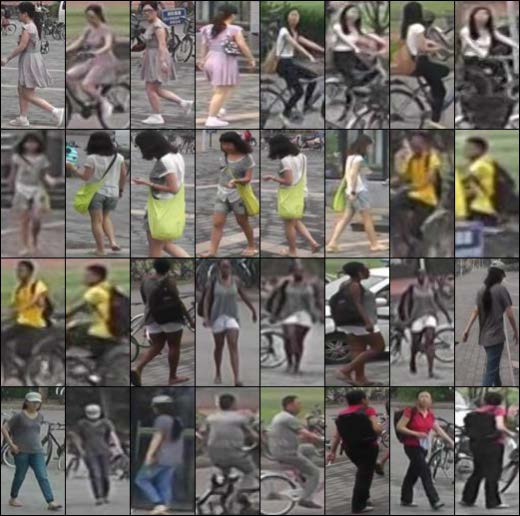}
  }%
  \subfigure[SSIM, Mis-ranking]{
		\includegraphics[width=0.32\textwidth]{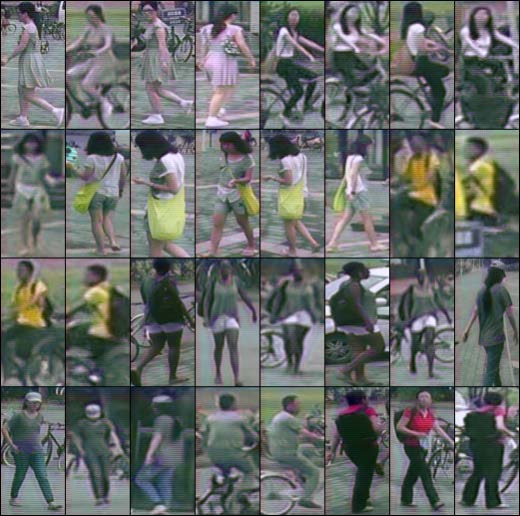}
  }%
  \subfigure[SSIM, LCYE]{
		\includegraphics[width=0.32\textwidth]{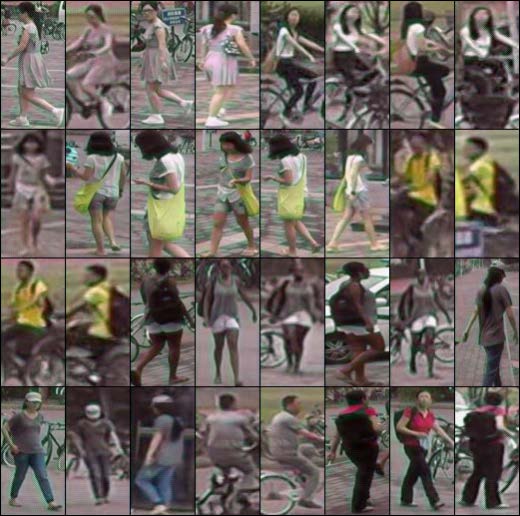}
  }%

  \subfigure[Original images (Same as (a))]{
		\includegraphics[width=0.32\textwidth]{figure6-original.jpg}
  }%
  \subfigure[MS-SSIM, Mis-ranking]{
		\includegraphics[width=0.32\textwidth]{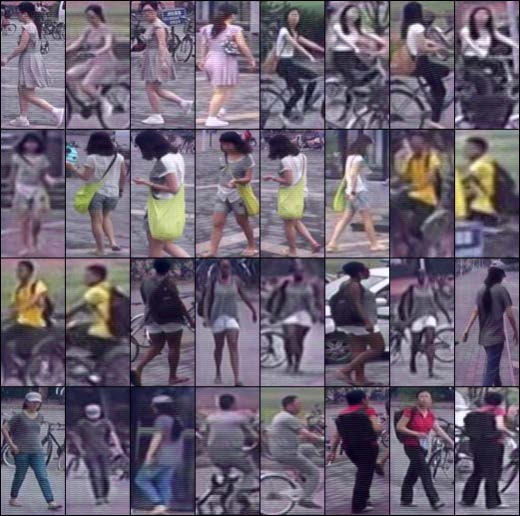}
  }%
  \subfigure[MS-SSIM, LCYE]{
		\includegraphics[width=0.32\textwidth]{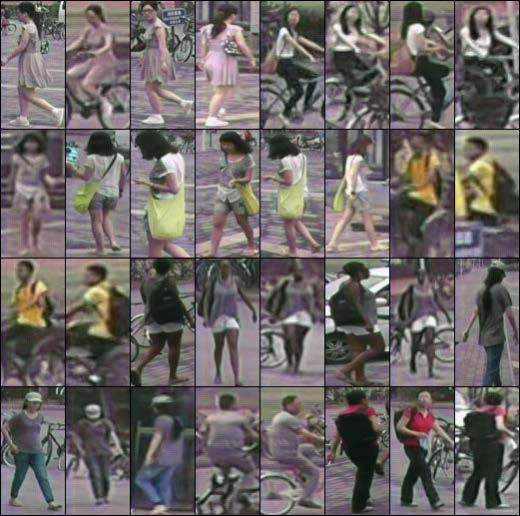}
  }%

  \caption{Visualization of using different visual perception losses for Mis-ranking and our method.}
  \label{figure vp}
\end{figure*}

\begin{table}[!t]
	\renewcommand{\arraystretch}{1.1}
	\caption{\textit{Target attack} results (\%) on Market1501.}
	\label{table black}
	\centering
	
	\begin{tabular}{c|c|cc}
		\hline
		\multirow{2}{*}{Method} & Target Consistency & \multicolumn{2}{c}{Attack} \\ \cline{2-4} 
		& Rank1 $\uparrow$             & Rank1  $\downarrow$       & mAP $\downarrow$       \\ \hline
		DenseNet121             &    71.2                &  2.0             & 1.3           \\
		AlignedReID             &   78.6                 &   4.2            & 1.7           \\
		SPGAN                   &     63.6               &    1.2           &    1.4        \\
		TransReID               &    45.0                &    2.1           &  2.0          \\ \hline
	\end{tabular}
\end{table}

\textbf{Target Attack.} To the best of our knowledge, our LCYE is the first method of target attack on ReID. In Table \ref{table black}, our method could successfully transfer the prototype of the desired identity to random images with over 60\% consistency accuracy and high attack performance, except on TransReID. One possible reason is that, compared to CNN, Transformers always focus on a larger area of recognition cues, making personalized target attacks harder. Moreover, compared with \textit{non-target attack}, \textit{i.e.,} white-box attack, the performance only drops 2\%, 4.1\%, 1.2\%, and 1.2\% for four models, respectively. It indicates our LCYE is not sensitive to personalized assignation.


\begin{figure*}[!t]
	\centering
	\subfigure[Original images]{
		\includegraphics[width=0.48\textwidth]{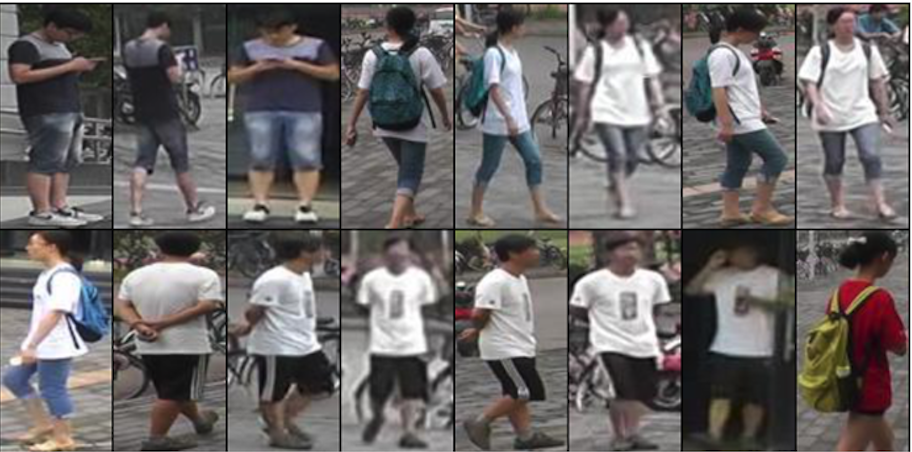}
	}%
	\subfigure[Original images (Same as (a))]{
		\includegraphics[width=0.48\textwidth]{figure7-original.png}
	}%
	
	\subfigure[Mis-ranking, $\varepsilon=3$]{
		\includegraphics[width=0.48\textwidth]{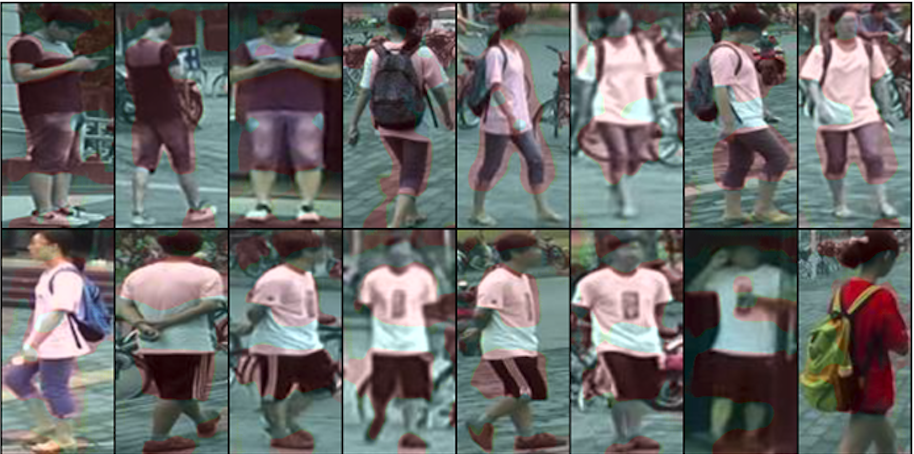}
	}%
	\subfigure[LCYE, $\varepsilon=3$]{
		\includegraphics[width=0.48\textwidth]{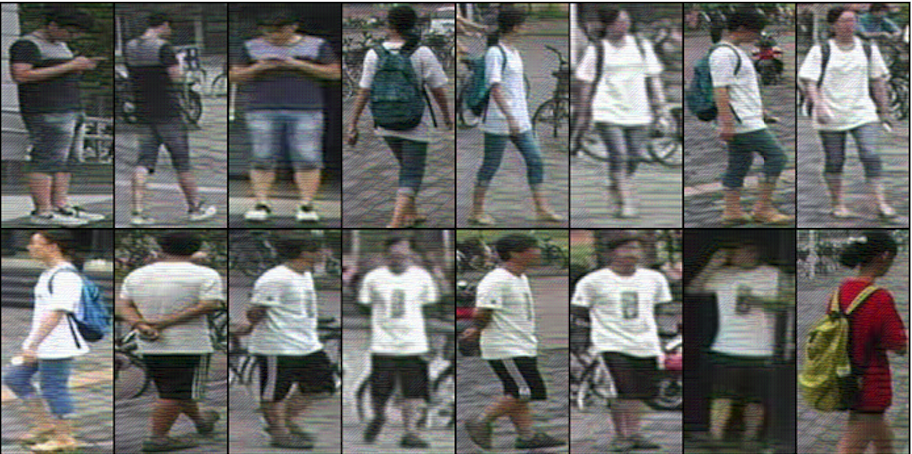}
	}%
	
	\subfigure[Mis-ranking, $\varepsilon=10$]{
		\includegraphics[width=0.48\textwidth]{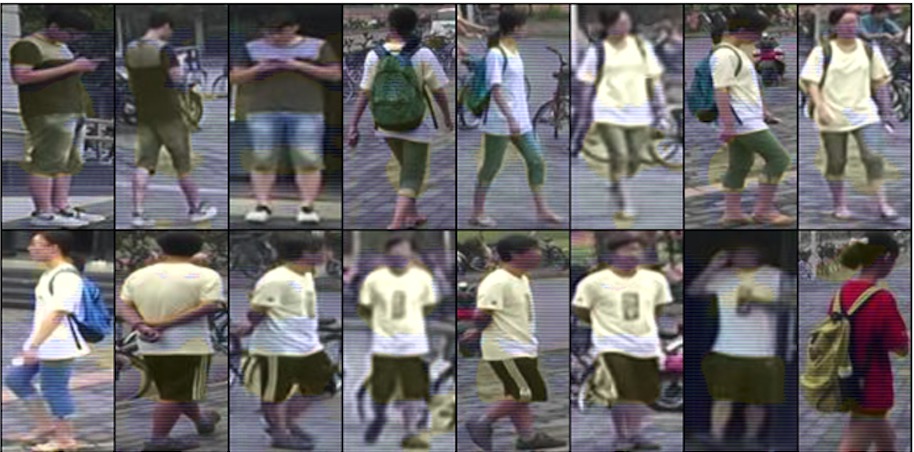}
	}%
	\subfigure[LCYE, $\varepsilon=10$]{
		\includegraphics[width=0.48\textwidth]{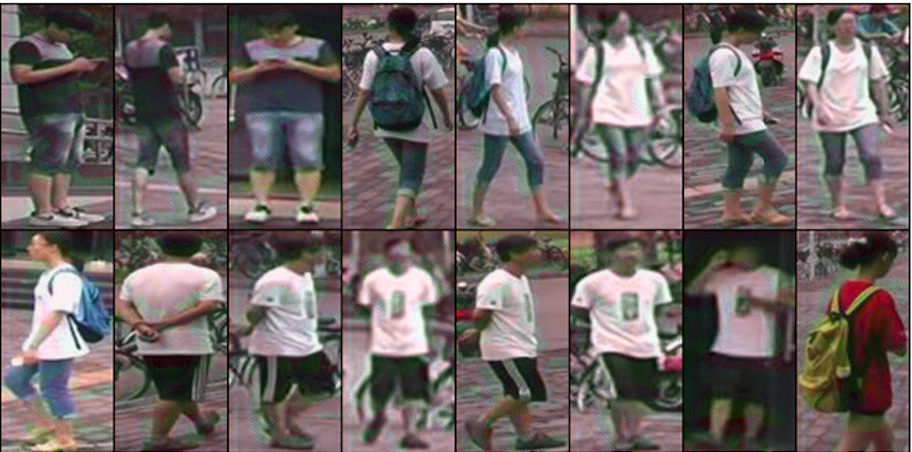}
	}%
	
	\subfigure[Mis-ranking, $\varepsilon=16$]{
		\includegraphics[width=0.48\textwidth]{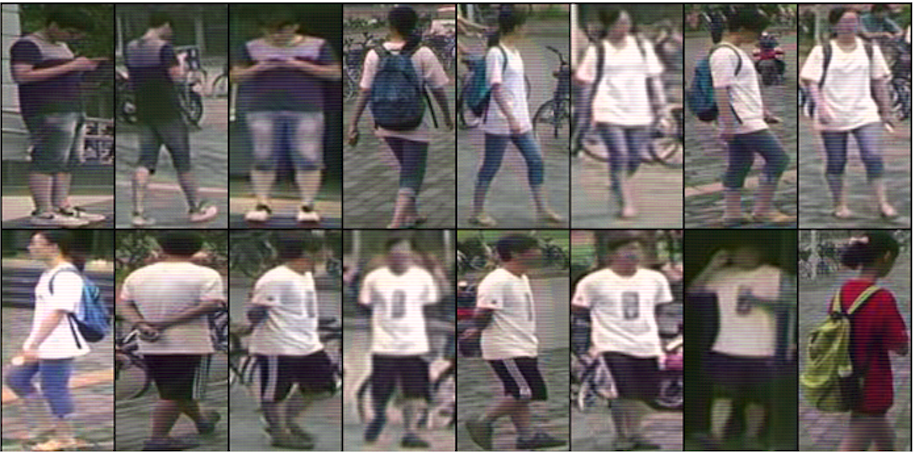}
	}%
	\subfigure[LCYE, $\varepsilon=16$]{
		\includegraphics[width=0.48\textwidth]{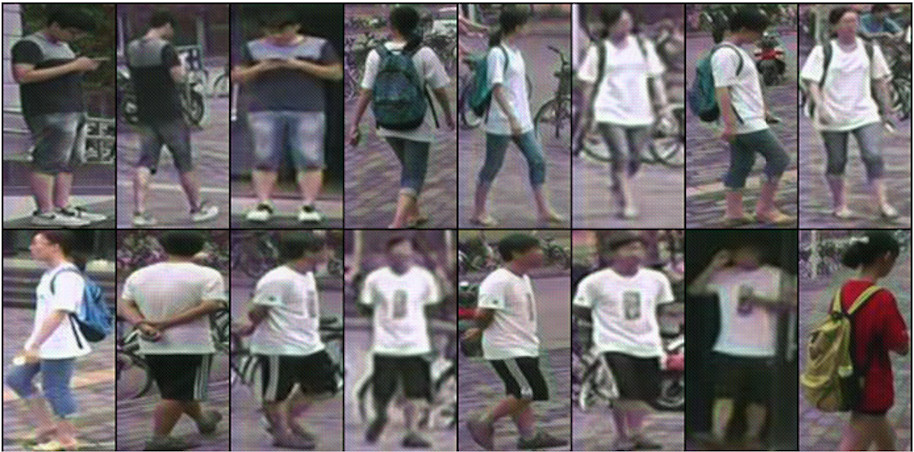}
	}%
	
	\caption{Visualization of using different visual perception losses for Mis-ranking and our method.}
	\label{figure var}
\end{figure*}

\begin{figure*}[!t]
  \centering
  \subfigure[Original images]{
		\includegraphics[width=0.48\textwidth]{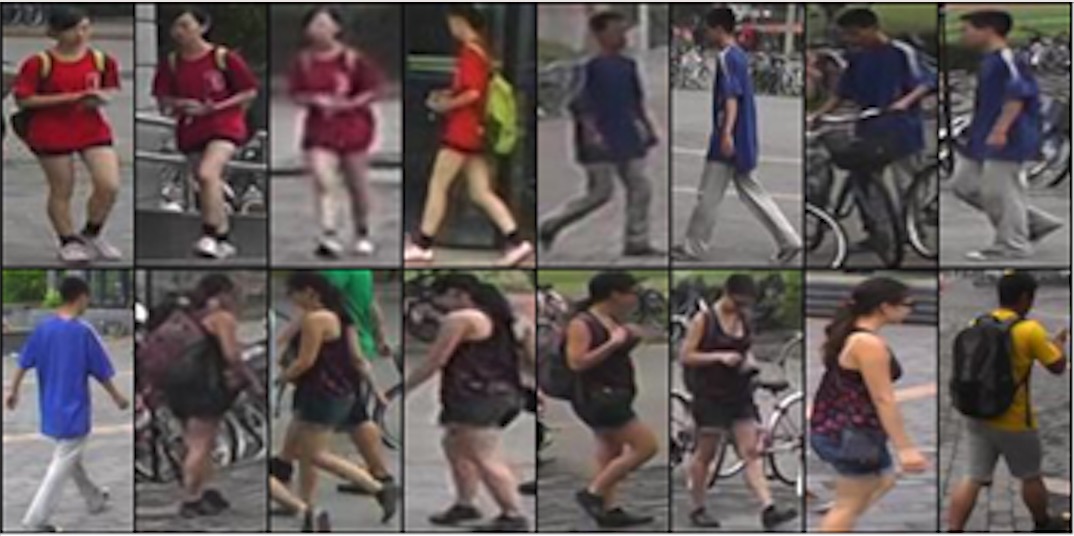}
  }%
  \subfigure[Original images (Same as (a))]{
		\includegraphics[width=0.48\textwidth]{figure8-original.png}
  }%
  
  \subfigure[Without Memory Module]{
		\includegraphics[width=0.48\textwidth]{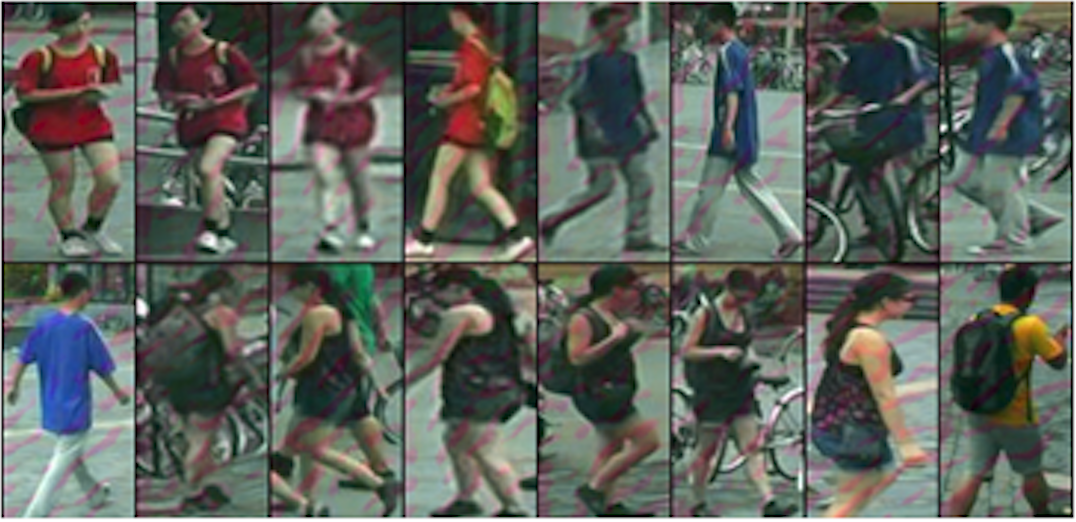}
  }%
  \subfigure[The Mask Map Without Memory Module]{
		\includegraphics[width=0.48\textwidth]{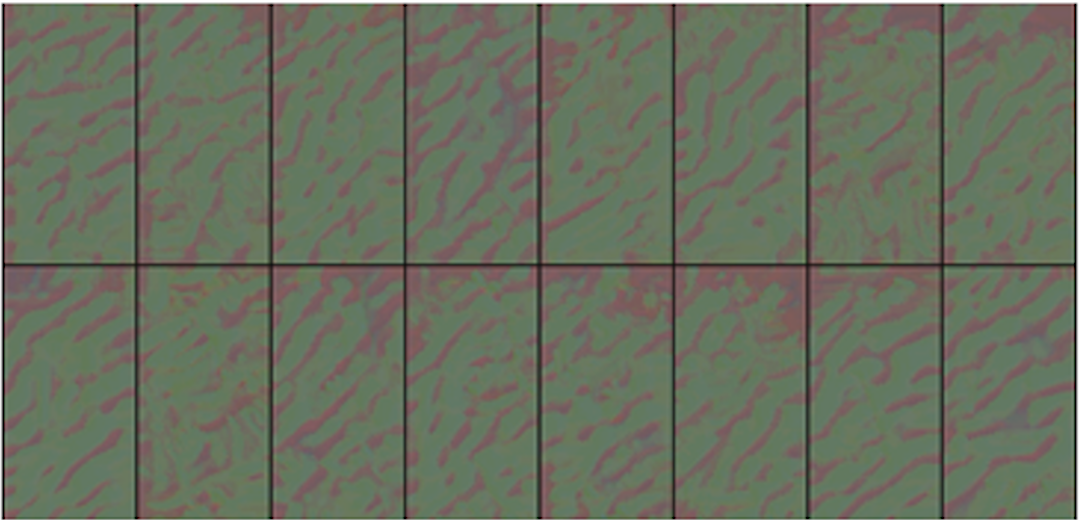}
  }%

  \subfigure[With Memory module]{
		\includegraphics[width=0.48\textwidth]{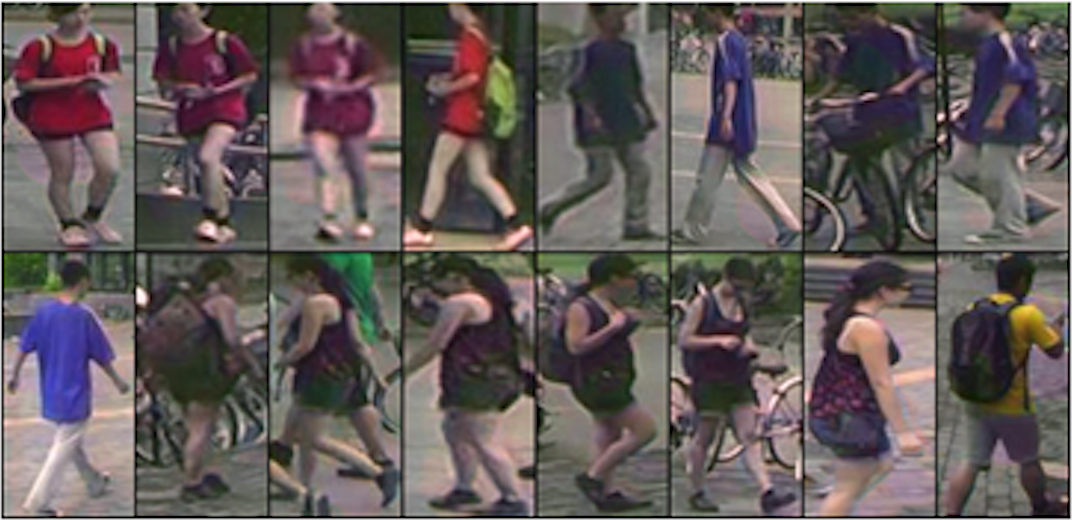}
  }%
  \subfigure[The Mask Map With Memory Module]{
		\includegraphics[width=0.48\textwidth]{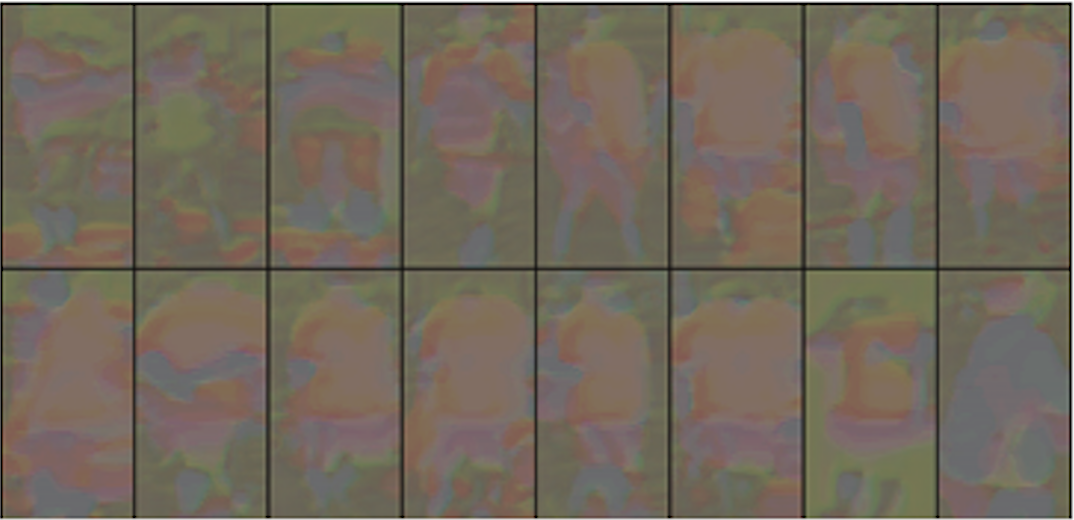}
  }%

  \caption{Visualization of our LCYE with/without memory module. And LCYE can clearly capture the contours of the human body from the mask map.}
  \label{figure cycle}
\end{figure*}

\subsection{Ablation Study}


\begin{table}[!t]
\renewcommand{\arraystretch}{1.1}
\caption{Ablation on `\textit{does the memory learn the belief of victim model?}' in Online Mimicking and Offline Mimicking with three mimicking variants. ReID results are evaluated on mimicking the branch.}
\label{table mimicking}
\centering
\begin{tabular}{c|cc|cc}
\hline
\multirow{2}{*}{Mimicking} & \multicolumn{2}{c|}{ReID} & \multicolumn{2}{c}{Attack} \\
                           & Rank1$\uparrow$        & mAP$\uparrow$       & Rank1$\downarrow$         & mAP $\downarrow$       \\ \hline
baseline                   & 61.3             &  54.2         &   5.5            &  2.1          \\
online                     &  85.2            & 74.3          &   5.6            &  2.1          \\
offline                    &   61.3           &  54.2         &   5.5            &  2.1          \\ \hline
\end{tabular}
\end{table}

\textbf{Cross-model \& Cross-domain Knowledge Distillation.} For cross-domain adaption and cross-model consensus, we use different $\mathcal{M}'$ to attack $\mathcal{M}$ with attack loss and find their attack performance is unexpectedly good without obvious distinction with Table \ref{table whitebox}. We believe that strong attack supervision may aggressively direct the whole optimization. Thus, we choose to remove the mis-ranking loss $\mathcal{L}_{mr}$ to check the model consensus for better interpretability. This experiment is similar to a black-box attack in meaning but free from attack guidance. As shown in Figure \ref{figure blackbox} (d)-(e), the diversity of knowledge consensus from models is more significant than from domains. Expect CUHK03, which also achieves the lowest value in cross-dataset attack; all domains' result seems uniform. This finding is also consistent with the cross-model dataset, indicating domain distribution makes fewer senses than a model structure for knowledge commonsense of robustness. As shown in Table \ref{table mimicking}, our LCYE achieves a similar performance as Table 1 in the original paper. For cross-domain knowledge distillation, it shows a similarly good performance. Thus, we believe the attack supervision may make whole optimization aggressive without showing the knowledge property of target model $\mathcal{M}$ and knowledge model $\mathcal{M}'$.

\textbf{Sensitivity of Visual Perception Loss.} As shown in Figure \ref{figure vp}, we provide more visualization to analyze the insensitivity of our method to different objectives. We can find that our method poses less dark green or purple background on original images and generates clearer images without heavy blur. We contribute this benefit to the interpolation of VM knowledge to both generator and discriminator since it also conveys the configuration of realistic images.


\textbf{Online Mimicking and Offline Mimicking.} One common concern is: Does the memory learn the belief of the victim model? One explanation is the cross-model/domain knowledge distillation and target attack, which show the specific knowledge of each model and the possibility of attack. Besides, to determine the influence of mimicking the manner in the interaction with the attacker, we conduct experiments with three variants, i.e., \textit{baseline mimicking, online mimicking, offline mimicking}, without attack loss. As shown in Table \ref{table mimicking}, ReID results evaluate the leftover property of the mimicking branch after our simple mimicking manner. The identical results of \textit{baseline} and \textit{offline} mean the memory module is not affected by the attacked branch during training and the quality of memory does not influence the performance of the attacker (joint training and two-stage training are different from memory retrieved by the attacker in each iteration). Furthermore, the obtained memory of \textit{online} is different from that of the victim model since the model structure is partly different. But the performance does not drop for this knowledge difference. We owe this phenomenon to the effectiveness of our LCYE paradigm, which is not sensitive to knowledge.

\begin{table}[!t]
\renewcommand{\arraystretch}{1.1}
\caption{Proportion of adversarial points.  $\dagger$ denotes the results with appropriate relaxation. The ratio denotes the adversarial points/total points.}
\label{table ratio}
\centering

\begin{tabular}{c|cc|cc}
\hline
\multirow{2}{*}{Ratio} & \multicolumn{2}{c|}{Mis-ranking} & \multicolumn{2}{c}{ours} \\ \cline{2-5} 
                       & Rank1           & mAP            & Rank1        & mAP       \\ \hline
full size              & 1.4             & 2.3            & 0.1          & 0.3       \\ \hline
1/2                    & 39.3            & 31.5           & 11.1         & 3.6       \\
1/4                    & 72.7            & 85.9           & 11.1         & 3.5       \\
1/8                    & 91.8            & 79.1           & 11.1         & 3.5       \\
1/16                   & 91.8            & 79.1           & 11.0         & 3.5       \\
1/16$\dagger$                   & 8.2             & 14.7           & 1.3          & 0.6       \\
1/32$\dagger$                   & 59.4            & 47.3           & 1.3          & 0.7       \\
1/64$\dagger$                  & 75.5            & 61.5           & 1.6          & 1.0       \\ \hline
\end{tabular}
\end{table}

\textbf{Number of the Pixels to be Attacked.} 
We further ablate the pixel demand of our method to attack in Table \ref{table ratio}. Our LCYE achieves promising performance even with a small ratio. Note that it keeps 11.1\% Rank1 and 3.6\% mAP from 1/2 to 1/16. We believe LCYE needs much fewer pixels than other attackers. The relaxation only brings about less than 10\% improvement, compared to Mis-ranking, which heavily depends on it. This benefit may come from the accurate hit of our mask predictor, which points out the salient region believed by the victim model.

\textbf{Comparisons of Different $\varepsilon$.}  Larger $\varepsilon$ could effectively boost attack performance but sacrifice the visual quality. We manually control the magnitude of $\varepsilon$ to verify the effectiveness of our LCYE. As shown in Table \ref{table var}, smaller $\varepsilon$ would not limit our attack performance. It meets the lower bound ($\backsim 0.4\%$) of Rank1 at $\varepsilon = 10$ or even much early. Especially when using $\varepsilon = 10$ or 5, our LCYE also shows promising superiority over Mis-ranking with 20\%-60\% gains. We further provide a visualization comparison in Figure \ref{figure var}. With small magnitudes of $\varepsilon$, Mis-ranking not only has strong blue atmospheres and obvious color blocks over original images but also poses striped Gaussian blur. However, our method generates a much more reasonable adversary compared to Mis-ranking.

\textbf{Cycle Consistency for Generation.} Our LCYE keeps cycle consistency via identity-aware adversarial learning and interpolating VM knowledge to generator and discriminator, which are termed as explicit and implicit guidance. Figure \ref{figure cycle} shows the noise and adversary visualization where the cycle consistency facilitates the inconspicuous perception.


\begin{table}[!t]
\renewcommand{\arraystretch}{1.1}
\caption{Ablation on different $\varepsilon$. The results is reported on AlignedReID with Market1501.}
\label{table var}
\centering
\resizebox{\linewidth}{!}{
\begin{tabular}{c|cccc|cccc}
\hline
\multirow{2}{*}{} & \multicolumn{4}{c|}{Mis-ranking} & \multicolumn{4}{c}{ours} \\ \cline{2-9} 
                  & R1     & R5     & R10    & mAP   & R1   & R5   & R10  & mAP \\ \hline
40                & 0.0    & 0.2    & 0.6    & 0.2   & 0.0  & 0.1  & 0.1  & 0.0 \\
20                & 0.1    & 0.4    & 0.8    & 0.4   & 0.0  & 0.1  & 0.1  & 0.0 \\
16                & 1.4    & 3.7    & 5.4    & 2.3   & 0.1  & 0.9  & 1.8  & 0.3 \\
10                & 24.4   & 38.5   & 46.6   & 21.0  & 0.4     & 1.6     &  3.2    & 0.3    \\
5                 & 69.2       & 82.6       &  87.0      & 56.4      &  4.2    &   10.2   &  15.1    & 1.6    \\
3                 &  83.9      &  92.5      & 95.1       &  70.2     & 8.1     &   18.1   & 23.8     & 2.7    \\ \hline
\end{tabular}}
\end{table}

\begin{table}[!t]
\renewcommand{\arraystretch}{1.2}
\caption{Complexity comparison with baseline model in training and testing.}
\centering
\resizebox{\linewidth}{!}{
\begin{tabular}{c|cc|cc}
\hline
     & \multicolumn{2}{c|}{Traning}         & \multicolumn{2}{c}{Testing}         \\ \hline
Model & \multicolumn{1}{c|}{baseline} & LCYE & \multicolumn{1}{c|}{baseline} & LCYE \\ \hline
Paras & \multicolumn{1}{c|}{2.5616 $\times$ $10^7$} & 3.0021  $\times$ $10^7$
& \multicolumn{1}{c|}{1.9508 $\times$ $10^7$ }   &  2.4733 $\times$ $10^7$   \\ \hline
FLOPs & \multicolumn{1}{c|}{1.9899 $\times$ $10^9$} & 3.6822 $\times$ $10^9$  & \multicolumn{1}{c|}{1.3614 $\times$ $10^9$}         &  2.8385 $\times$ $10^9$    \\ \hline
\end{tabular}}
\label{Model Complex}
\end{table}

\textbf{\textit{Discussion}.} The transferability of ReID models are always evaluated by applying it to another dataset where the uninspiring results usually show the poor generalization ability of models. We instead analyze the commonsense of the vulnerability of different models/domains and find that even without attack guidance, domain-specific and model-specific decision-making is fragile, with a minor distribution shift coming from the knowledge belonging to others or even themselves. It means the improvement brought from either data augmentation (GAN) or local awareness (Transformer) can not facilitate the robustness across models and domains in robustness. We reconsider the capability of our LCYE, including lower pixel demand and lower objective dependency, from the aspect of inverse reinforcement learning. Namely, the attacker estimates a possible distribution of adversaries by the attack award from victim models. Our LCYE benefits from (1) a more direct way to observe and memorize the environment (VM) and (2) providing dense awards for each pixel. In particular, the memory module is a shortcut interface accessing the decision-making process of the victim model rather than indirectly predicting it by attack award. As shown in Figure \ref{figure decision}, we illustrate the decision-making from the image level in neurons. Our method redirects the map from each salient locality to the same wrong higher semantics since it assigns the same identity prototype to the image, while current methods always scatter them without pixel-level supervision.


\begin{figure}[!t]
  \centering
  \subfigure[Mis-ranking with different attack losses]{
		\includegraphics[width=0.8\linewidth]{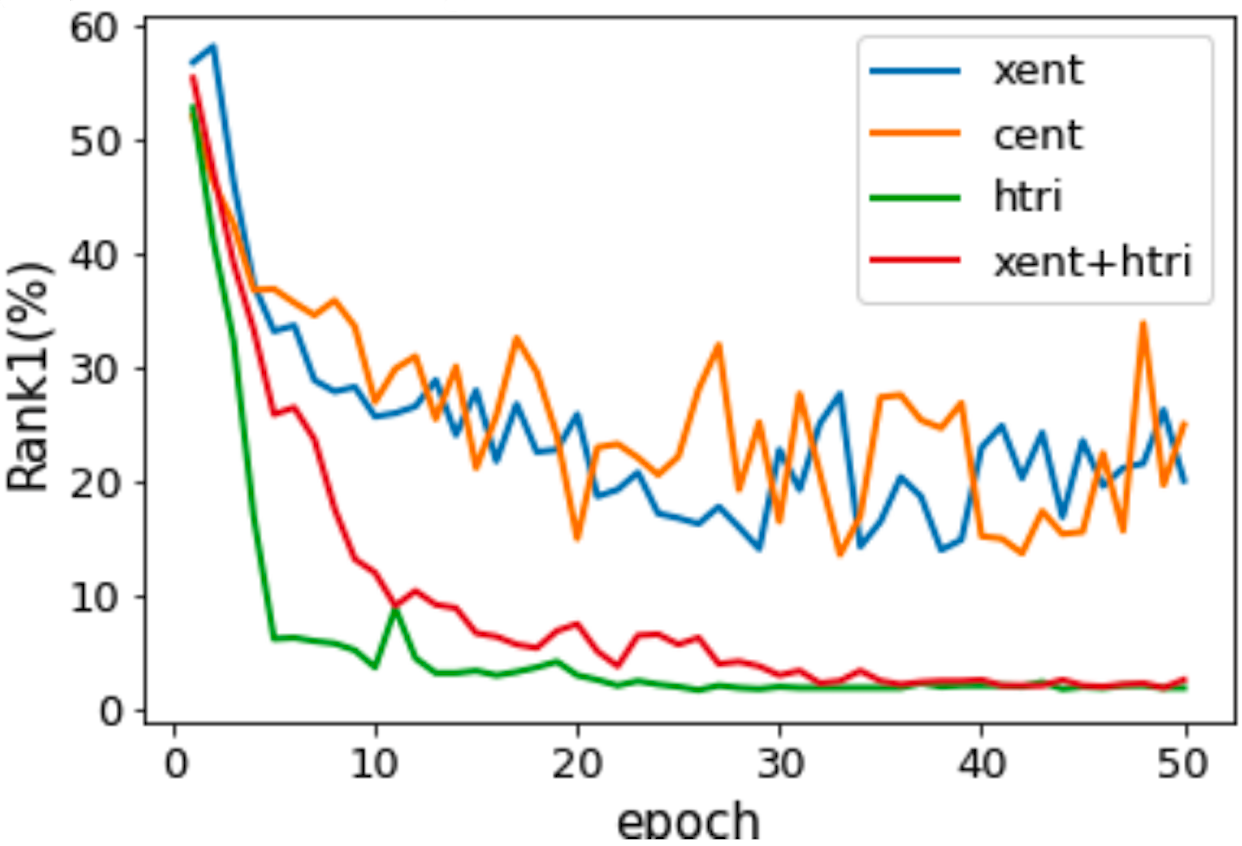}
  }%
  
  \subfigure[LCYE with different attack losses]{
		\includegraphics[width=0.8\linewidth]{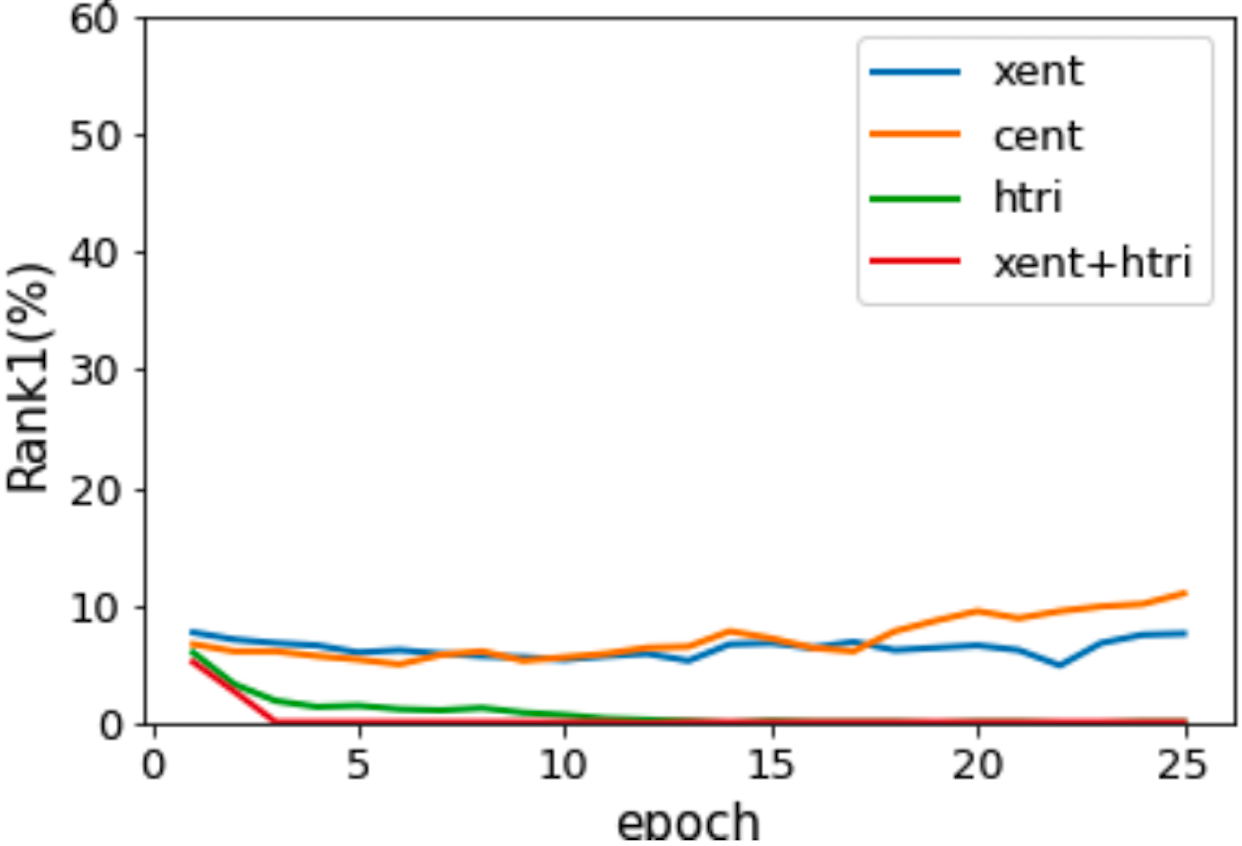}
  }%

  \caption{Ablation on the sensitivity of different losses. (a) and (b) record the test evaluation results after each epoch.}
  \label{figure loss}
\end{figure}


\textbf{Sensitivity of Different Losses.} Mis-ranking heavily relies on attack loss $\mathcal{L}_{mr}\in \{\textbf{cent, xent,etri, xcent+etri}\}$ and $\mathcal{L}_{VP}\in \{\textbf{SSIM, MS-SSIM}\}$, since it needs enough adversarial feedback via the overturning victim model. In Figure \ref{figure loss}, our LCYE has sharper and more stable curves than Mis-ranking using different attack losses, indicating that memory reduces the necessity of well-designed objectives. This observation is also consistent with Figure \ref{figure blackbox}, where our LCYE achieves promising attack performance without attack supervision. Moreover, we also ablate the dependence on visual perception loss. As shown in Figure \ref{figure blackbox}, our method has a more inconspicuous visualization than the counterpart without dark purple or green background. Thus, exquisite losses are not a necessity of our LCYE.
\label{loss_sensity}

\textbf{Complexity Comparison.} As shown in Table \ref{Model Complex}, our LCYE could boost the performance by a large margin with minor complexity addition.  The additional parameter complexity mainly comes from the prototype memory module in both training and testing. Besides, the FLOPs complexity is primarily increased in training since the victim model needs to process clean images for mimicking branches, while in testing, we directly use the memory module for attack without processing images in the victim model again.

\section{Conclusion and Future Work}
\label{sec6}
In this paper, we propose our LCYE to attack victim models by cheating their minds. Unlike current attackers using the standard attack paradigm, our LCYE boosts the accessibility of victim models' decision-making process through our teacher-student mimicking. We also conduct extensive experiments on the knowledge commonsense of vulnerability across domains and models. Our LCYE is a transferable attacker with promising performance on white-box, black-box, and target attacks. We tackle the robustness attack and generalization attack jointly without further refined manipulation. However, currently, generalization capabilities like domain adaptation and domain generalization, are discussed independently, which is inversely represented as the out-of-distribution non-transferable attack in this case. For example, AlignedReID trained on Market1501 could also achieve promising results on CUHK03 and DukeMTMC. An out-of-distribution non-transferable attack could influence its performance on CUHK03 while never hurting it on Market1501 and DukeMTMC. This conditional and directional attack is much more dangerous than the standard one and could further connect model  interpretability from positive (proxy tasks like classification and retrieve) and negative (adversarial attack) aspects.  We believe future work could focus on this topic, with the help of explicit victim model knowledge.

\section*{Acknowledgements}
This work was supported in part by the Guangdong Key Lab of AI and Multi-modal Data Processing, BNU-HKBU United International College (UIC) under Grant No. 2020KSYS007 and Computer Science Grant No. UICR0400025-21; the National Natural Science Foundation of China (NSFC) under Grant No. 61872239 and No. 62202055; the Institute of Artificial Intelligence and Future Networks, Beijing Normal University; the Zhuhai Science-Tech Innovation Bureau under Grants No. ZH22017001210119PWC and No. 28712217900001; and the Interdisciplinary Intelligence Supercomputer Center of Beijing Normal University (Zhuhai).

\ifCLASSOPTIONcaptionsoff
  \newpage
\fi

\bibliographystyle{IEEEtran}
\bibliography{reference}

\begin{IEEEbiography}[{\includegraphics[width=1in,height=1.25in,clip,keepaspectratio]{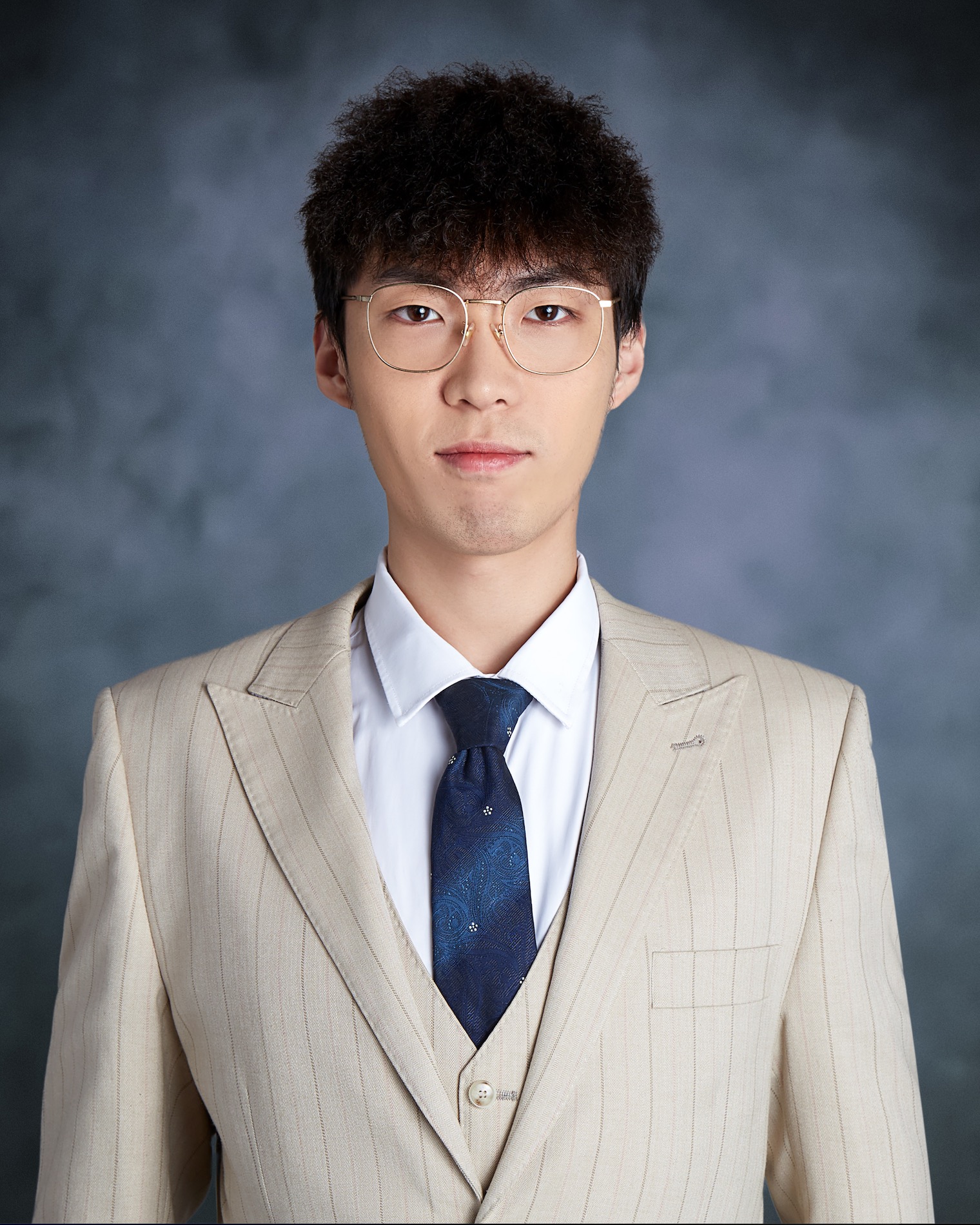}}]{Mingjie Wang}
    received the B.S. degree from the Department of Computer Science and Technology, Longdong University, China, in 2021 and is currently an M.Phil. candidate in the Department of Data Science and Technology, BNU-HKBU United International College (UIC). His current research interests include Time Series Analysis, NLP, Machine Learning, and Deep Learning.
\end{IEEEbiography}

\begin{IEEEbiography}[{\includegraphics[width=1in,height=1.25in,clip,keepaspectratio]{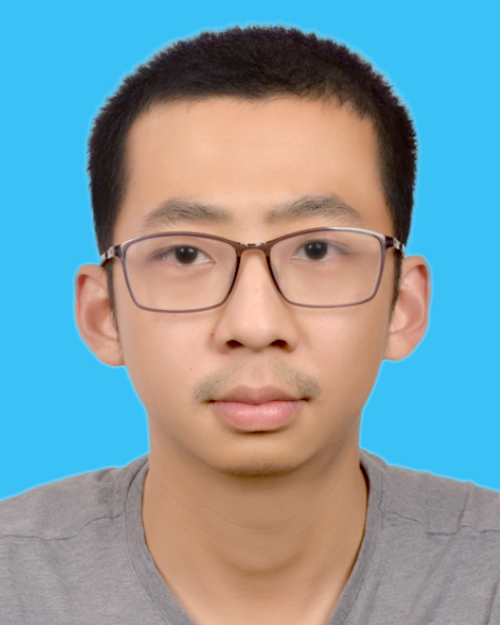}}]{Jianxiong Guo}
	received his Ph.D. degree from the Department of Computer Science, University of Texas at Dallas, Richardson, TX, USA, in 2021, and his B.E. degree from the School of Chemistry and Chemical Engineering, South China University of Technology, Guangzhou, China, in 2015. He is currently an Assistant Professor with the Advanced Institute of Natural Sciences, Beijing Normal University, and also with the Guangdong Key Lab of AI and Multi-Modal Data Processing, BNU-HKBU United International College, Zhuhai, China. He is a member of IEEE/ACM/CCF. He has published more than 40 peer-reviewed papers and been the reviewer for many famous international journals/conferences. His research interests include social networks, wireless sensor networks, combinatorial optimization, and machine learning.
\end{IEEEbiography}

\begin{IEEEbiography}[{\includegraphics[width=1in,height=1.25in,clip,keepaspectratio]{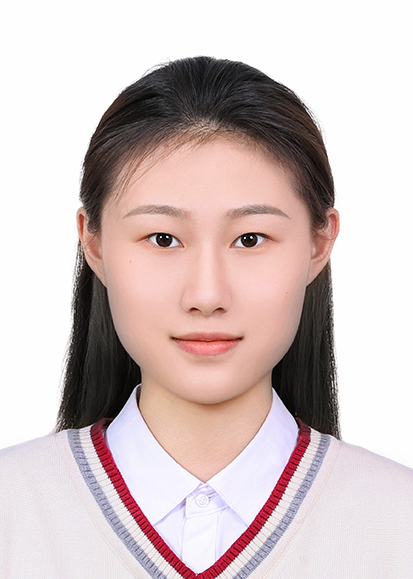}}]{Sirui Li} received the B.S. degree from the Department of Statistics and Data Science, BNU-HKBU United International College (UIC), in 2023 and is currently an M.Sc. candidate in Halıcıoğlu Data Science Institute, University of California San Diego. Her main interests are Data Mining, Image Analysis, and Interpretable Deep Learning.
\end{IEEEbiography}

\begin{IEEEbiography}[{\includegraphics[width=1in,height=1.25in,clip,keepaspectratio]{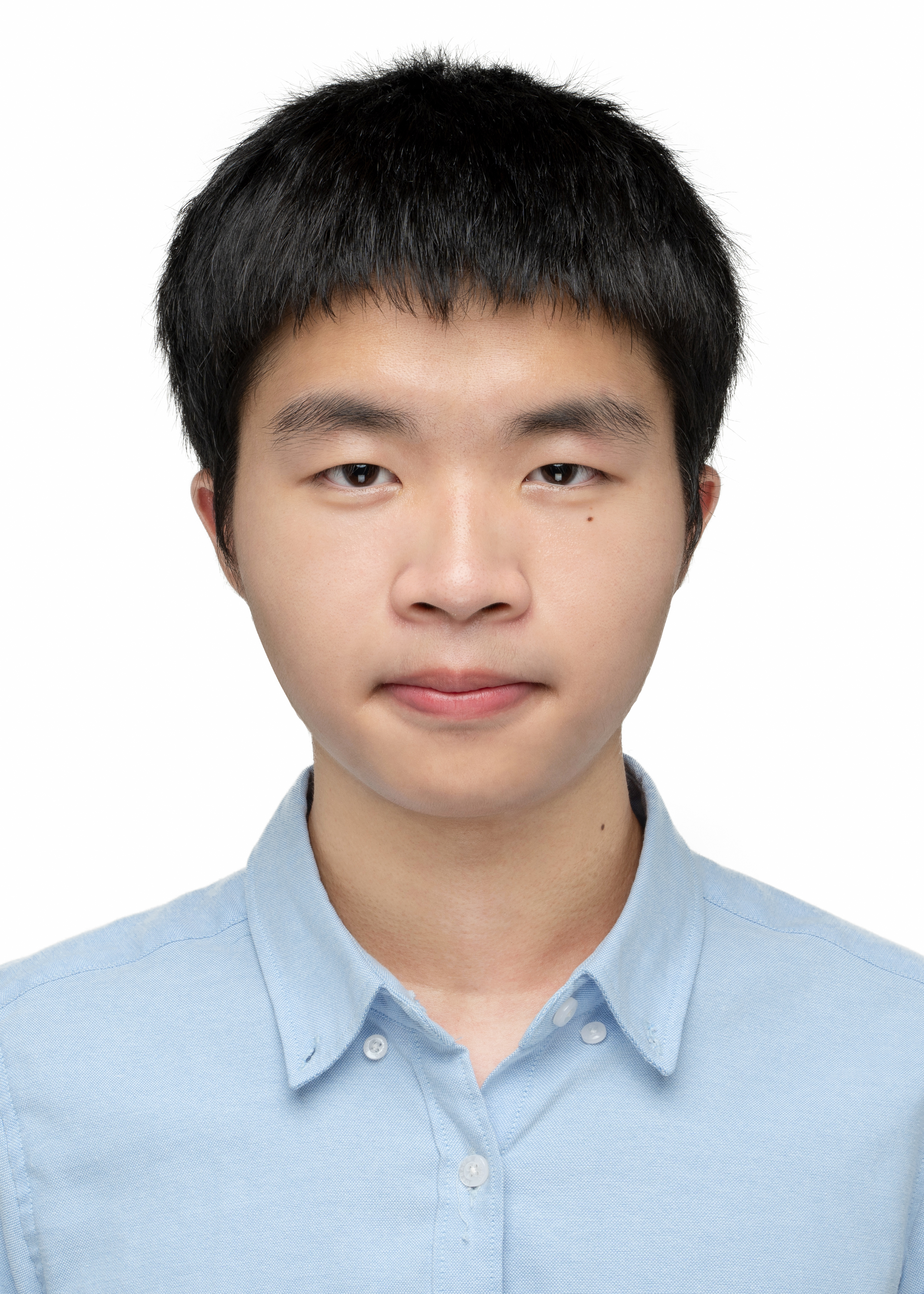}}]{Dingwen Xiao} received the B.S. degree from the Department of Statistics and Data Science, BNU-HKBU United International College (UIC), in 2023 and is currently an M.Sc. candidate in Data Driven Modeling, Hong Kong University of Science and Technology. His main interests are Data Mining, Image Semantic Analysis, and Interpretable Deep Learning.
\end{IEEEbiography}

\begin{IEEEbiography}[{\includegraphics[width=1in,height=1.25in,clip,keepaspectratio]{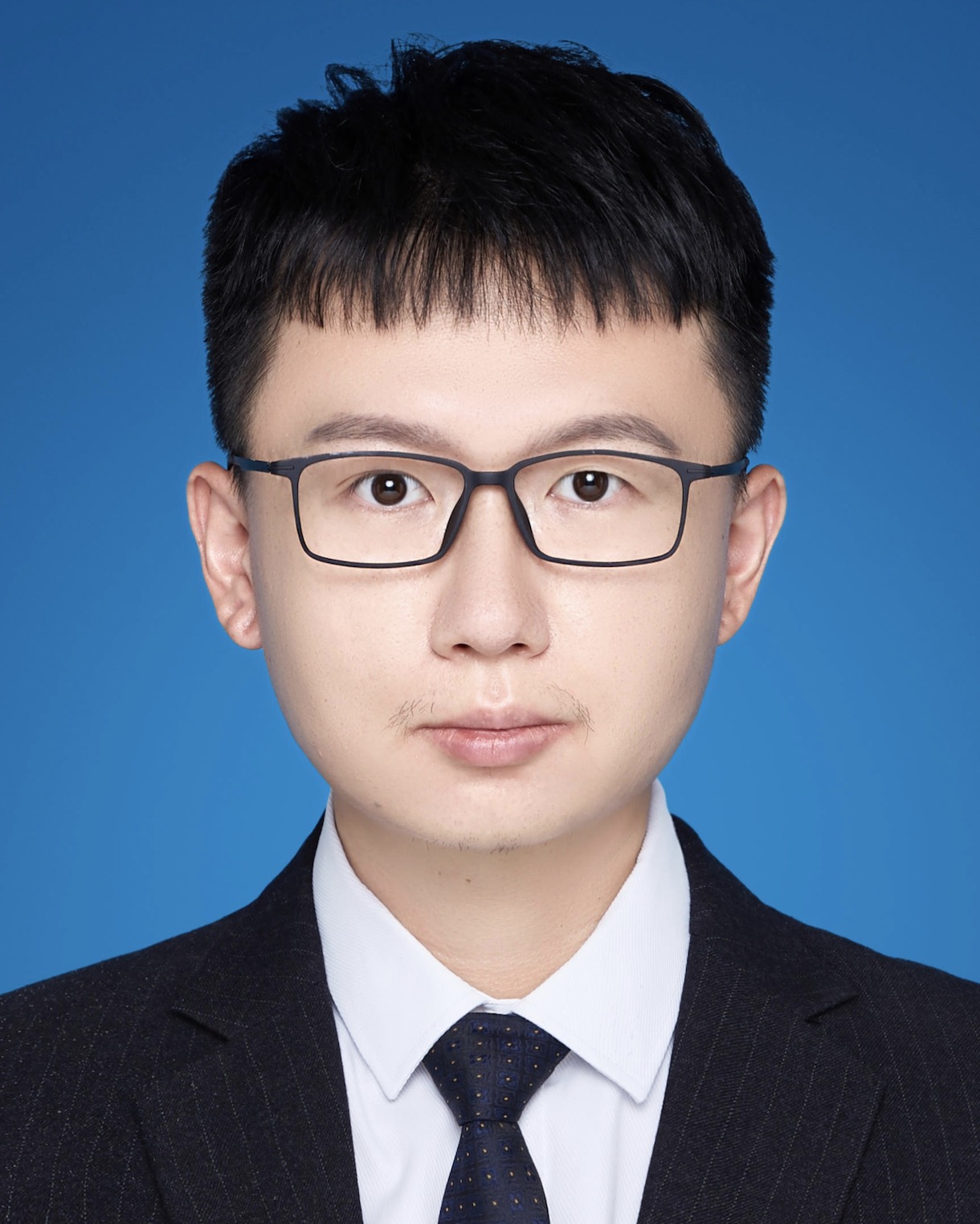}}]{Zhiqing Tang}
    received the B.S. degree from School of Communication and Information Engineering, University of Electronic Science and Technology of China, China, in 2015 and the Ph.D. degree from Department of Computer Science and Engineering, Shanghai Jiao Tong University, China, in 2022. He is currently an assistant professor with the Advanced Institute of Natural Sciences, Beijing Normal University, China. His current research interests include edge computing, resource scheduling, and reinforcement learning.
\end{IEEEbiography}

\end{document}